%% file: main.tex
\documentclass[10pt,journal,compsoc]{IEEEtran}
\usepackage{graphicx}
\usepackage{capt-of}
\usepackage{amsmath,amssymb,amsfonts,dsfont,bm,bbm,mathrsfs,pifont}
\usepackage[ruled,linesnumbered]{algorithm2e}
\usepackage{listings}
\usepackage{booktabs,multirow,adjustbox}
\usepackage{float}
\usepackage[table]{xcolor}
\definecolor{citeblue}{HTML}{0071bc}
\usepackage[pagebackref,breaklinks,colorlinks,citecolor=citeblue]{hyperref}
\usepackage[capitalize]{cleveref}  
\usepackage{wrapfig}
\crefname{section}{Sec.}{Secs.}
\Crefname{section}{Section}{Sections}
\crefname{table}{Tab.}{Tabs.}
\Crefname{table}{Table}{Tables}
\crefname{figure}{Fig.}{Figs.}
\Crefname{figure}{Figure}{Figures}
\crefname{equation}{Eq.}{Eqs.}
\Crefname{equation}{Equation}{Equations}
\crefname{algorithm}{Algorithm}{Algorithms}
\definecolor{textpurple}{RGB}{135,89,201}

\renewcommand{\paragraph}[1]{\vspace{1.25mm}\noindent\textbf{#1}}
\newcommand{\method}{HPM\xspace}

\definecolor{codegreen}{rgb}{0,0.6,0}
\definecolor{codegray}{rgb}{0.5,0.5,0.5}
\definecolor{codepurple}{rgb}{0.58,0,0.82}
\definecolor{backcolour}{rgb}{1.0,1.0,1.0}
\lstdefinestyle{mystyle}{
    backgroundcolor=\color{backcolour},
    commentstyle=\color{codegreen},
    keywordstyle=\color{magenta},
    numberstyle=\tiny\color{codegray},
    stringstyle=\color{codepurple},
    basicstyle=\ttfamily\scriptsize,
    breakatwhitespace=false,
    breaklines=true,
    captionpos=b,
    keepspaces=true,
    showspaces=false,
    showstringspaces=false,
    showtabs=false,
    tabsize=2
}
\definecolor{textpurple}{RGB}{135,89,201}

\definecolor{upcolor}{RGB}{57,182,74}
\newcommand{\up}[1]{\textcolor{upcolor}{$\uparrow$ #1}}
\newcommand{\down}[1]{\textcolor{red}{$\downarrow$ #1}}
\definecolor{deemph}{gray}{0.75}
\newcommand{\gc}[1]{\textcolor{deemph}{#1}}
\newcommand{\pub}[1]{\tiny\textcolor{deemph}{[#1]}}
\newlength\savewidth\newcommand\shline{\noalign{\global\savewidth\arrayrulewidth
  \global\arrayrulewidth 1pt}\hline\noalign{\global\arrayrulewidth\savewidth}}
\definecolor{Light}{rgb}{0.99, 0.92, 0.95}

\ifCLASSOPTIONcompsoc
  \usepackage[nocompress]{cite}
\else
  \usepackage{cite}
\fi

\begin{document}

\title{Bootstrap Masked Visual Modeling via \\ Hard Patches Mining}
%
%
%
%

\author{Haochen~Wang,
        Junsong~Fan,
        Yuxi~Wang,
        Kaiyou~Song,
        Tiancai~Wang,
        Xiangyu~Zhang,
        and~Zhaoxiang~Zhang$^*$,~\IEEEmembership{Senior~Member,~IEEE}
\IEEEcompsocitemizethanks{
\IEEEcompsocthanksitem 
H. Wang, J. Fan, Y. Wang, and Z. Zhang are with the Center for Research on Intelligent Perception and Computing, State Key Laboratory of Multimodal Artificial Intelligence Systems, Institute of Automation, Chinese Academy of Sciences, Beijing 100190, China. 
E-mail: \{wanghaochen2022, junsong.fan, zhaoxiang.zhang\}@ia.ac.cn, yuxiwang93@gmail.com
\IEEEcompsocthanksitem
H. Wang and Z. Zhang are also with the University of Chinese Academy of Sciences, Beijing 100190, China.
\IEEEcompsocthanksitem 
J. Fan, Y. Wang, and Z. Zhang are also with the Centre for Artificial Intelligence and Robotics, Hong Kong Institute of Science \& Innovation, Chinese Academy of Sciences, Hong Kong 999077, China.
\IEEEcompsocthanksitem
K.Song, T. Wang, and X. Zhang are with Megvii Technology, Beijing 100096, China. Email: songkaiyou@foxmail.com, \{wangtiancai, zhangxiangyu\}@megvii.com
}
\thanks{$^*$Corresponding author.}
}

\input{sections/0.abs}

\maketitle

\IEEEdisplaynontitleabstractindextext

%
\IEEEpeerreviewmaketitle

\input{sections/1.intro}

\input{sections/2.related}

\input{sections/3.method}

\input{sections/4.experiments}

\input{sections/5.conclusion}

\ifCLASSOPTIONcaptionsoff
  \newpage
\fi



\bibliographystyle{IEEEtran}
\bibliography{IEEEabrv,ref.bib}
%

%








\end{document}

%% file: sections/0.abs.tex
\IEEEtitleabstractindextext{%
\begin{abstract}
Masked visual modeling has attracted much attention due to its promising potential in learning generalizable representations.
Typical approaches urge models to predict specific contents of masked tokens, which can be intuitively considered as teaching a \textit{student} (the model) to solve given problems (predicting masked contents).
Under such settings, the performance is highly correlated with mask strategies (the difficulty of provided problems).
%
We argue that it is equally important for the model to stand in the shoes of a \textit{teacher} to produce challenging problems by itself.
Intuitively, patches with high values of reconstruction loss can be regarded as hard samples, and masking those hard patches naturally becomes a demanding reconstruction task.
To empower the model as a teacher, we propose Hard Patches Mining (\method), predicting patch-wise losses and subsequently determining where to mask.
Technically, we introduce an auxiliary loss predictor, which is trained with a relative objective to prevent overfitting to exact loss values.
%
%
Also, to gradually guide the training procedure, we propose an easy-to-hard mask strategy.
Empirically, \method brings significant improvements under both image and video benchmarks.
%
%
%
Interestingly, solely incorporating the extra loss prediction objective leads to better representations, verifying the efficacy of determining where is hard to reconstruct.
The code is available at \url{https://github.com/Haochen-Wang409/HPM}.
\end{abstract}


}

%% file: sections/1.intro.tex
\IEEEraisesectionheading{
\section{Introduction}\label{sec:intro}}

\IEEEPARstart{S}{elf}-supervised learning~\cite{he2020momentum, chen2020simple, grill2020bootstrap, chen2021empirical, chen2021exploring, wang2023hard, wang2023droppos}, with the goal of learning generalizable feature representations from large-scale datasets without any human annotations, has become a burgeoning research focal point in computer vision (CV).
Inspired by masked language modeling (MLM)~\cite{devlin2018bert, radford2018improving, radford2019language, brown2020language} in natural language processing (NLP), where the model is urged to predict masked words within a sentence, masked visual modeling (MVM), a parallel context within CV, has attracted numerous interests of researchers to learn scalable \textit{image} representations~\cite{bao2021beit, he2022masked, xie2022simmim} and \textit{video} representations~\cite{feichtenhofer2022masked, tong2022videomae}.

\begin{figure}[t]
    \centering
    \includegraphics[width=1\linewidth]{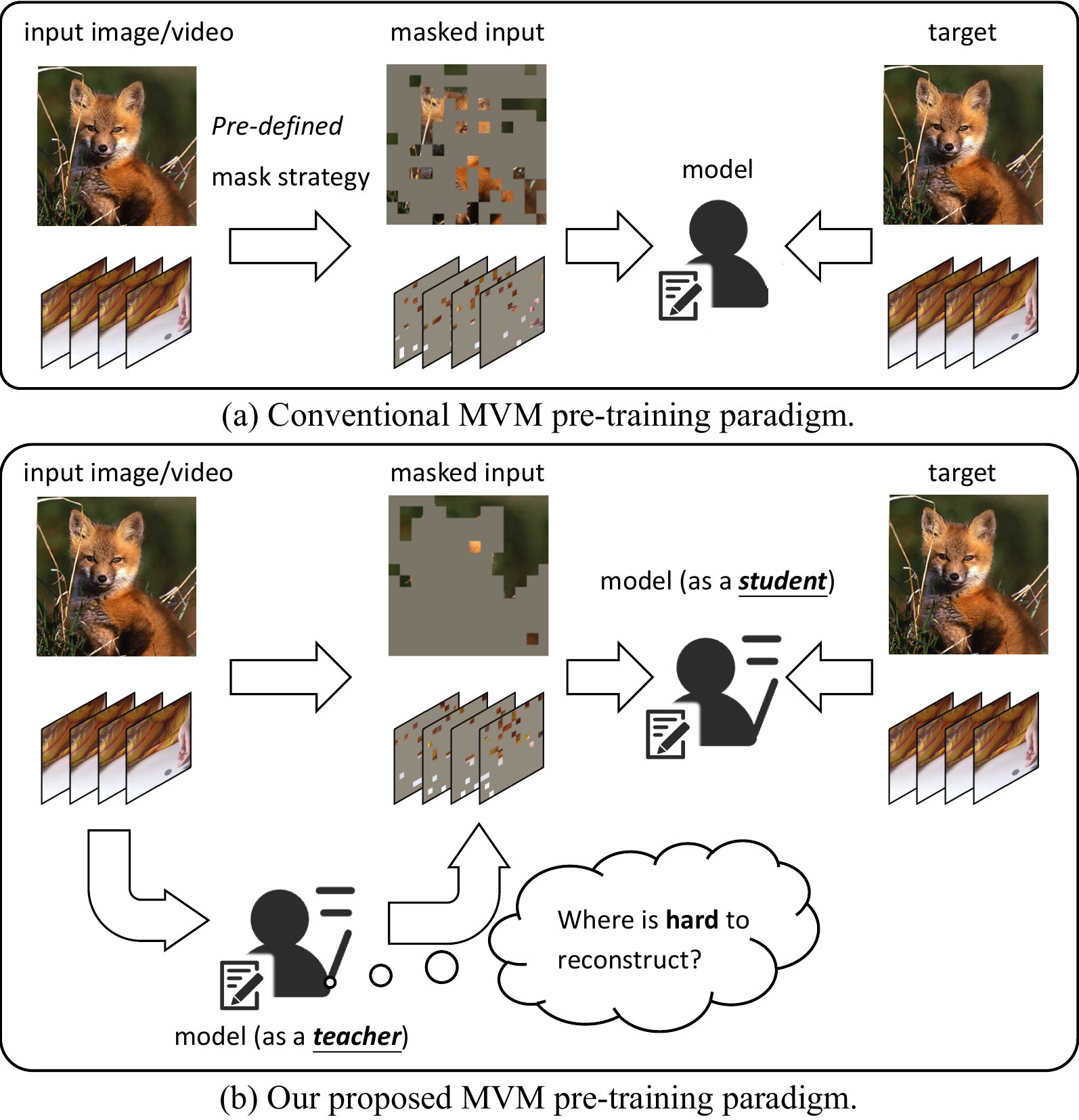}
    \vspace{-20pt}
    \caption{
    Comparison between conventional MVM paradigm and our proposed \method.
    \textbf{(a)} Conventional approaches can be interpreted as training a \textit{\textbf{student}}, where the model is \textit{only} equipped with the ability to solve a given problem under some pre-defined mask strategies.
    \textbf{(b)} Our proposed \method pre-training paradigm makes the model to be both a \textbf{\textit{teacher}} and a \textbf{\textit{student}}, with the extra ability to \textit{produce a challenging pretext task}.
    This design allows iteratively bootstrapping the outputs of the network in an alternative way for enhanced representations.
    }
    \label{fig:motivation}
    \vspace{-10pt}
\end{figure}

\begin{figure*}[t]
    \centering
    \includegraphics[width=1\linewidth]{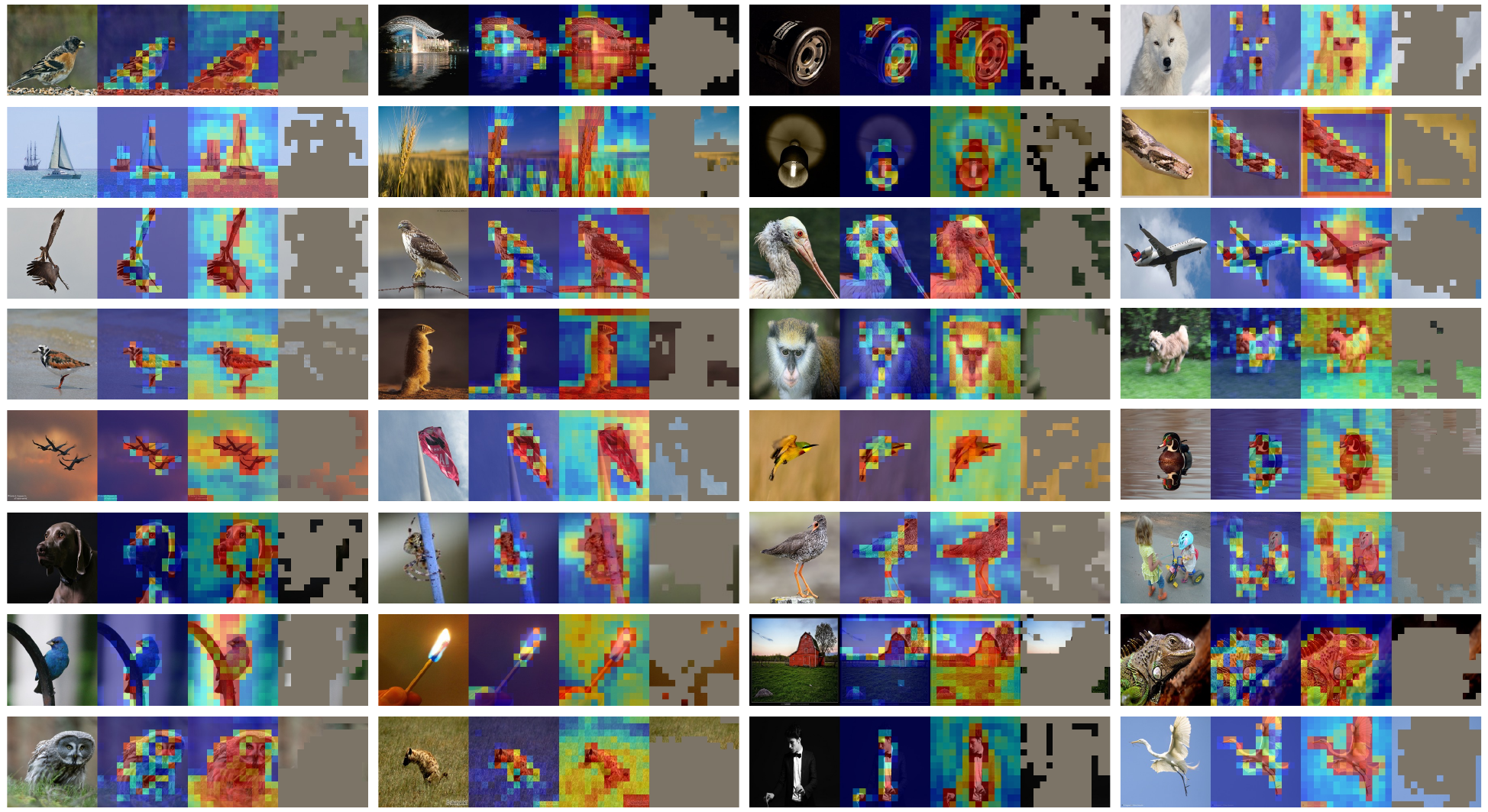}
    \vspace{-20pt}
    \caption{
    Visual comparison between \textbf{reconstruction loss} and \textbf{discriminativeness} on ImageNet \textit{validation} set.
    We load the pre-trained ViT-B/16~\cite{dosovitskiy2020image} provided by MAE~\cite{he2022masked}.
    For each tuple, we show the \textbf{(a)} \textit{input image}, \textbf{(b)} \textit{patch-wise reconstruction loss} averaged over 10 different masks, \textbf{(c)} \textit{predicted loss}, and \textbf{(d)} \textit{masked images} generated by the predicted loss (\textit{i.e.}, patches with top 75\% predicted loss are masked).
    Red means higher loss while blue indicates the opposite.
    \textit{Discriminative parts tend to be hard to reconstruct}.
    }
    \label{fig:discriminative}
\end{figure*}

\cref{fig:motivation}\textcolor{red}{a} illustrates the paradigm of conventional approaches for MVM pre-training.
In these typical solutions~\cite{he2022masked, xie2022simmim, bao2021beit, tong2022videomae, feichtenhofer2022masked}, models are trained to predict specific contents of masked patches.
%
%
To make this process more challenging and alleviate the spatial redundancy, the design of mask strategies becomes crucial~\cite{wang2023hard, he2022masked}.
Usually, masks are generated under pre-defined manners, \textit{e.g.}, random masking~\cite{he2022masked, feichtenhofer2022masked, xie2022simmim}, block-wise masking~\cite{bao2021beit}, uniform masking~\cite{li2022uniform}, tube masking~\cite{tong2022videomae}, and even motion-guided masking~\cite{huang2023mgmae, fan2023motion}.
%
%
Intuitively, the procedure shown in \cref{fig:motivation}\textcolor{red}{a} can be considered as training a \textit{\textbf{student}} (the model) on solving visual problems (predicting specific masked contents).
%
%
However, we argue that focusing on solving problems like a student is far from enough.
It is equally important to \textit{learn how to create these challenging problems}, thus taking on the roles of both a \textit{\textbf{student}} and a \textit{\textbf{teacher}}.
Illustrated in \cref{fig:motivation}\textcolor{red}{b}, by learning to create challenging problems and solving them \textit{simultaneously}, the model is forced to hold a more comprehensive understanding of the visual contents, and thus iteratively leads itself by generating more difficult tasks and solving them.
%

To this end, we propose \textit{Hard Patches Mining (\method)}, a new training paradigm for MVM.
Specifically, given input visual signals, instead of generating a binary mask under pre-defined criteria, the model first acts as a teacher, identifying hard patches and subsequently producing challenging problems and then, as a student, predicts masked patches.
%
%
Then, the question becomes (1) how to measure the difficulty of each patch, and (2) how to make the model aware of where the hard patches are.
If the above two issues are well addressed, simply masking those hard patches detected by the model becomes a solution.

\textit{First}, to measure patch-wise difficulties, we observe that \textit{the reconstruction loss} naturally becomes a metric of the difficulty of the MVM task\footnote{For visualization, we use the official checkpoint from \url{https://dl.fbaipublicfiles.com/mae/visualize/mae\_visualize\_vit\_base.pth}} (see the first two elements of each tuple in \cref{fig:discriminative}).
%
%
Therefore, patches with larger losses (red patches in \cref{fig:discriminative}) can be regarded as hard samples for reconstruction. 
Interestingly, those hard patches turn out to be discriminative parts of an image (\textit{e.g.}, object).
\textit{Second}, to make the model aware of hard patches, we encourage it to \textit{predict} reconstruction loss for each patch.
Consequently, with the capability of detecting hard patches, masking those patches with higher predicted losses naturally contributes to demanding MVM tasks.
Technically, we introduce an auxiliary loss predictor, predicting patch-wise losses first and deciding where to mask next based on its outputs.
To prevent it from being overwhelmed by the exact values of reconstruction losses and make it concentrate on the \textit{relative} relationship among patches, we present a novel objective.

Qualitative results of the loss predictor are provided in \cref{fig:discriminative}, where a ViT-B/16~\cite{dosovitskiy2020image} pre-trained with only 200 epochs on ImageNet-1K~\cite{russakovsky2015imagenet} is used for visualization.
As the third element for each tuple suggests, patches with larger \textit{predicted} losses tend to be discriminative, and thus masking these patches brings a challenging situation.
However, as illustrated in the last element for each tuple in \cref{fig:discriminative}, directly using the mask generated by predicted losses means we force the model to reconstruct forehead \textit{with almost only background}, which makes no sense.
%
To this end, we come up with an easy-to-hard mask generation strategy, providing some reasonable hints for reconstruction. 

Empirically, we observe significant and consistent improvements over the supervised baseline and vanilla MVM pre-training under \textit{both} image benchmarks and video benchmarks.
Concretely, under image SSL benchmarks, with only 800 epochs pre-training, \method achieves 84.2\% and 85.8\% Top-1 accuracy on ImageNet-1K~\cite{russakovsky2015imagenet} using ViT-B and ViT-L, outperforming MAE~\cite{he2022masked} pre-trained with 1600 epochs by +0.6\% and +0.7\%, respectively.
We further evaluate transfer learning on semantic segmentation, where \method achieves better results than supervised pre-training and other self-supervised counterparts.
Under video SSL benchmarks, with 800 epochs of pre-training, \method achieves 69.8\% and 80.8\% Top-1 accuracies on Something-Something v2~\cite{goyal2017something} and Kinetics-400~\cite{kay2017kinetics}, respectively, outperforming the original VideoMAE~\cite{huang2023mgmae} with +0.5\% and +0.8\%.
With longer pre-training schedules, \textit{i.e.}, 2400 epochs on Something-Something v2~\cite{goyal2017something} and 1600 epochs on Kinetics-400~\cite{kay2017kinetics}, \method manages to surpass state-of-the-art alternatives consistently.

\paragraph{Extension of the conference version.}
This paper extends our conference version~\cite{wang2023hard} in the following aspects.
First, we successfully extend our conference version~\cite{wang2023hard} from the image domain to the video domain, indicating that masking hard patches when pre-training is beneficial to \textit{both} domains.
Second, we empirically show that \method brings consistent and significant improvements over baselines for masked video modeling with \textit{minimal} domain knowledge and \textit{few} inductive biases.
The extra redundancy in the temporal dimension makes little impact for \method and \textit{almost no modification is necessary} for \method when adapting \method from the image domain to the video domain.
%
%
The only modification is that the mask of each frame is generated independently.
The success of \method in both domains demonstrates that both (1) difficult pre-training tasks and (2) the ability to propose challenging problems are crucial in masked visual modeling.
We hope our findings will inspire future works in designing a more appropriate masked visual modeling training paradigm.

%% file: sections/2.related.tex
\section{Related Work}\label{sec:related}

In this section, we provide a brief introduction to the related literature, including self-supervised learning and masked visual modeling.
Furthermore, a detailed discussion on mask strategies in masked visual modeling is provided.

\subsection{Self-Supervised Visual Representation Learning}
The common goal of self-supervised learning (SSL) approaches in computer vision is to learn scalable and generalizable visual representations without any human annotations.
In the literature, how to design an appropriate pretext task, \textit{i.e.}, producing appropriate supervision with \textit{only} visual signals, becomes the crux.

The design of \textit{image} pre-text tasks has been widely explored.
Prior arts proposed to predict the relative location of a pair of image patches~\cite{doersch2015unsupervised}, solve jigsaw puzzles~\cite{noroozi2016unsupervised}, or colorize the given images~\cite{zhang2016colorful}.
In the past five years, instance discrimination~\cite{wu2018unsupervised} became popular because, for the first time, the performances of self-supervised algorithms surpassed the supervised counterpart on a wide range of downstream benchmarks~\cite{he2020momentum, oord2018representation, grill2020bootstrap}.
The core idea of these methods is to urge the model to learn view-invariant features, and thus they strongly depend on carefully designed data augmentation techniques~\cite{chen2020simple, caron2021emerging}.
Several works~\cite{wang2022semi, wang2023pulling, wang2023using} even extended this idea to semantic segmentation for a more discriminative representation space.
Very recently, Wang \textit{et al.}~\cite{wang2023droppos} proposed an interesting pre-text task, \textit{i.e.}, reconstructing dropped positions, for pre-training vision transformers.
The procedure is simple: they first drop a large random subset of positional embeddings and then urge the model to classify the actual position for each non-overlapping image patch among all possible positions.

Motivated by the success in the image domain, there has been great interest in exploring appropriate SSL pre-text tasks for \textit{videos}.
Some early works focused on extending image SSL methods for videos without special consideration in the temporal dimension~\cite{ahsan2019video, huo2020self, kim2019self}.
Other works, on the opposite, focused on capturing temporal coherence, such as future prediction~\cite{srivastava2015unsupervised, lotter2016deep, mathieu2016deep, vondrick2016anticipating, walker2016uncertain, vondrick2018tracking}, object motion~\cite{agrawal2015learning, wang2015unsupervised, pathak2017learning, wang2019learning}, temporal ordering~\cite{misra2016shuffle, fernando2017self, lee2017unsupervised, wei2018learning, xu2019self}, and spatiotemporal contrast~\cite{han2019video, feichtenhofer2021large, qian2021spatiotemporal, recasens2021broaden}.

Very recently, masked visual modeling (MVM) has become very popular because of its superior performance and generalization ability under both image~\cite{bao2021beit, he2022masked} and video~\cite{feichtenhofer2022masked, tong2022videomae} benchmarks.
Therefore, this paper mainly focuses on MVM.
A detailed discussion of related literature is provided as follows.

\subsection{Masked Visual Modeling}
With the goal of building a unified self-supervised pre-training framework towards general artificial intelligence, MVM has attracted numerous interests of many researchers~\cite{bao2021beit, he2022masked, wei2022masked, zhou2021ibot, xie2022simmim, wang2023hard}, because of the great success of MLM~\cite{devlin2018bert, radford2018improving, radford2019language, brown2020language} and its autoregressive variants in NLP.
It is believed that the success of MVM paves a path that SSL in vision \textit{``may now be embarking on a similar trajectory as in NLP''}~\cite{he2022masked}.

Specifically, MVM is a specific paradigm of predictive learning.
It predicts any unobserved (masked) part from the observed part of the given signal~\cite{gupta2023siamese}.
A brief introduction is provided as follows.

\paragraph{Masked image modeling.}
Early attempts either predicted masked spatial patches~\cite{vincent2008extracting, vincent2010stacked, pathak2016context} or color channels~\cite{larsson2016learning, zhang2016colorful, zhang2017split, larsson2017colorization} based on visible ones.
In these methods, an autoencoder~\cite{hinton2006reducing} reconstructed input images after the information passed through a low-dimensional bottleneck layer.
Recently, masked image modeling approaches trained a Vision Transformer (\textit{e.g.}, ViT~\cite{dosovitskiy2020image} or its hierarchical variants~\cite{liu2021swin, wang2021pyramid, liu2022swin}) to predict specific contents of those masked patches.
The choice of reconstruction targets became a key design.
Specifically, BEiT~\cite{bao2021beit} predicted discrete tokens generated by a pre-trained dVAE~\cite{rolfe2016discrete}, and many other works~\cite{he2022masked, xie2022simmim, liu2022mixmim, li2022uniform} simply adopt raw RGB pixels as reconstruction targets.
MaskFeat~\cite{wei2022masked} even explored HoG~\cite{dalal2005histograms} features to be the target.
Also, frequency~\cite{liu2022devil, xie2022masked}, and features from a momentum teacher~\cite{zhou2021ibot, baevski2022data2vec, yi2022masked, wu2022extreme, dong2022bootstrapped} were widely exploited.

\paragraph{Masked video modeling.} 
By treating time as an extra isotropic dimension of the spatial dimension, masked image modeling methods can be easily extended to masked video modeling to learn high-quality spatiotemporal representations, as VideoMAE~\cite{tong2022videomae} and MAEST~\cite{feichtenhofer2022masked} did.
VideoMAE v2~\cite{wang2023videomae} further scaled up those models by introducing a dual masking strategy.
UMT~\cite{li2023unmasked} distilled knowledge from an unmasked vision foundation model, \textit{i.e.}, CLIP~\cite{radford2021learning}, contributing to a more efficient training procedure.
OmniMAE~\cite{girdhar2023omnimae} extended this paradigm to a unified pre-training of image and video modalities and outperformed models trained for a single modality.
However, some works argued that the temporal dimension is special and symmetrically treating spatial and temporal information might be sub-optimal.
To this end, MaskViT~\cite{gupta2022maskvit} and SiamMAE~\cite{gupta2023siamese} investigated predicting the masked future frame based on the visible current frame.
With a similar motivation, Salehi \textit{et al.}~\cite{salehi2023time} came up with a novel time-tuning paradigm.

\begin{figure*}[t]
    \centering
    \includegraphics[width=0.85\linewidth]{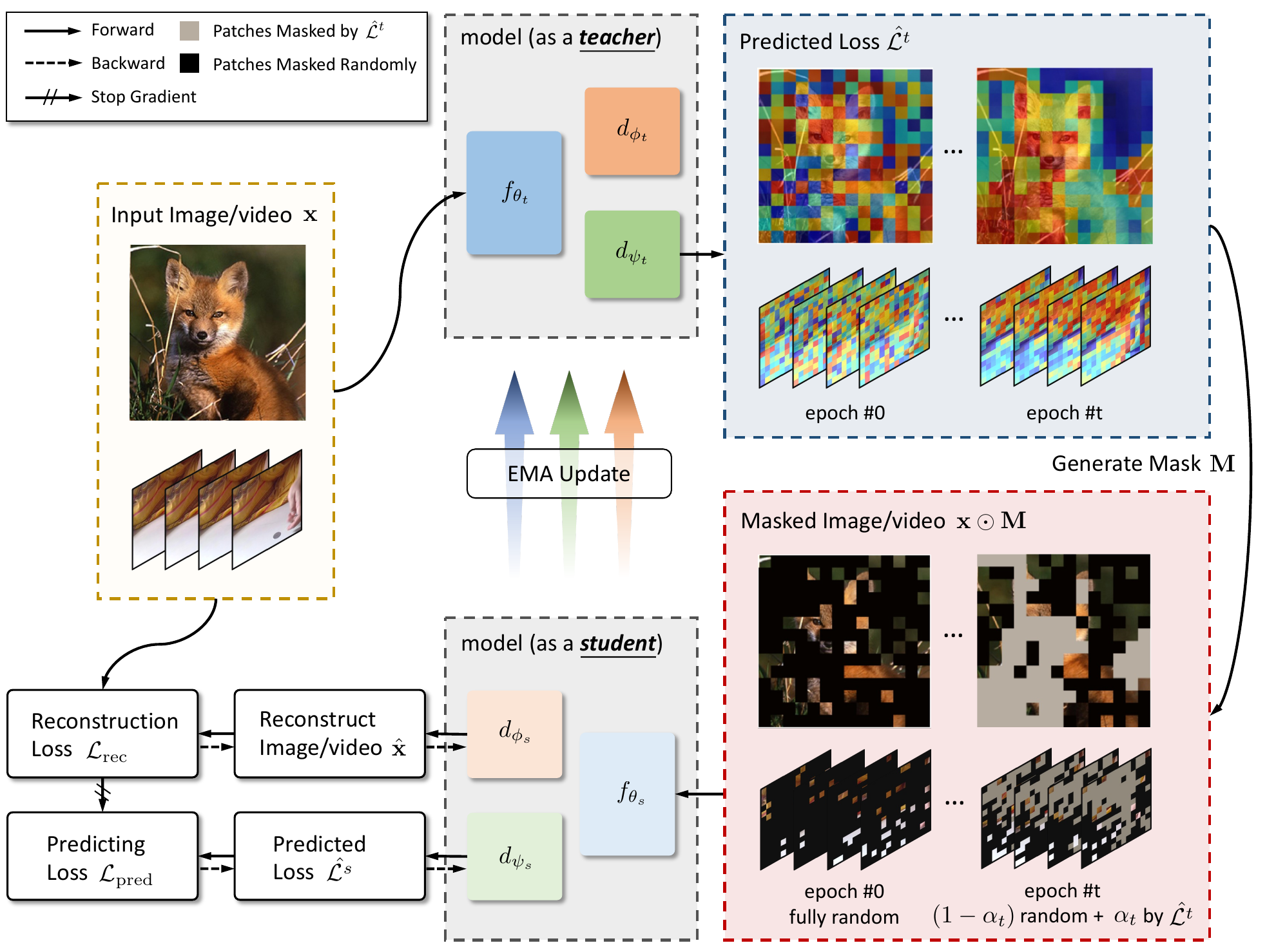}
    \vspace{-10pt}
    \caption{
    \textbf{Illustration of our proposed \method},
    containing a student network and a teacher network, where the teacher is updated by the student in an exponential moving average (EMA) manner.
    Each network consists of an encoder $f_{\theta}$, an visual reconstructer $d_{\phi}$, and a loss predictor $d_{\psi}$, parameterized by $\theta$, $\phi$, and $\psi$, respectively.
    For each input during pre-training, it is first fed into the teacher to predict the patch-wise reconstruction loss.
    Then, a binary mask is generated based on the current epoch and the predicted loss.
    Finally, only visible patches are fed into the student to 1) reconstruct masked patches defined in \cref{eq:mim}, and 2) predict relative loss defined in \cref{eq:bce}.
    %
    }
    \label{fig:pipeline}
\end{figure*}

\subsection{Mask Strategies for Masked Visual Modeling}

The success of masked visual modeling heavily relies on a carefully designed mask strategy due to the information redundancy~\cite{he2022masked, tong2022videomae}.
%
%
In the following, a detailed discussion of mask strategies is provided.

\paragraph{Mask strategies for masked image modeling.}
In the image domain, to address the spatial redundancy, MAE~\cite{he2022masked} used a large mask ratio (\textit{i.e.}, 75\%), BEiT~\cite{bao2021beit} adopt block-wise masking, and SimMIM~\cite{xie2022simmim} found that larger mask kernels (\textit{e.g.}, 32$\times$32) are more robust against different mask ratios.
Furthermore, to produce a more challenging pretext task, AttMask~\cite{kakogeorgiou2022hide} masked those patches with high attention signals.
ADIOS~\cite{shi2022adversarial} employed adversarial learning and trained an extra U-Net~\cite{ronneberger2015u} based masking model.
SemMAE~\cite{li2022semmae} regarded semantic parts as the visual analog of words and trained an extra StyleGAN~\cite{karras2019style} based decoder distilled by iBOT~\cite{zhou2021ibot}.
UM-MAE~\cite{li2022uniform} masked one patch in each 2$\times$2 local window, enabling pyramid-based ViTs (\textit{e.g.}, PVT~\cite{wang2021pyramid}, CoaT~\cite{xu2021co}, and Swin~\cite{liu2021swin, liu2022swin}) to take the random sequence of partial vision tokens as input.

\paragraph{Mask strategies for masked video modeling.}
An appropriate mask strategy even plays a more important role in masked video modeling because of the extra temporal redundancy.
To address this issue, both MAEST~\cite{feichtenhofer2022masked} and VideoMAE~\cite{tong2022videomae} used an extremely high masking ratio, \textit{i.e.}, 90\%.
Also, to relieve the information leakage in the temporal dimension, VideoMAE~\cite{tong2022videomae} further proposed the tube mask strategy, \textit{i.e.}, the binary mask shares across all frames.
However, this simple and straightforward tube masking approach implicitly assumes almost no motion between adjacent frames.
To this end, several approaches explicitly incorporated motion information to reduce temporal information leakage~\cite{huang2023mgmae, fan2023motion, ahmadian2023mofo}.
Specifically, to capture motion information and build temporally consistent masking volume, MGMAE~\cite{huang2023mgmae} leveraged an extra online optical flow estimator~\cite{teed2020raft}.
Similarly, MGM~\cite{fan2023motion} exploited the H.264 codec~\cite{richardson2002video} which obtains cheap motion correspondence to generate spatiotemporally continuous 3D masks.
MOFO~\cite{ahmadian2023mofo} proposed a motion-guided fine-tuning paradigm to further boost the performance.
Motivated by the policy gradient algorithm in reinforcement learning~\cite{kaelbling1996reinforcement}, AdaMAE~\cite{bandara2023adamae} masked more tokens from regions with high spatiotemporal information.

All the masking models of these methods are either pre-defined~\cite{he2022masked, wei2022masked, xie2022simmim, bao2021beit, zhou2021ibot, baevski2022data2vec, kakogeorgiou2022hide, huang2023mgmae, tong2022videomae, fan2023motion, ahmadian2023mofo} or separately learned~\cite{shi2022adversarial, li2022semmae}.
However, we argue that \textit{learn to mask} the discriminative parts is crucial, which can not only guide the model in a more challenging manner, but also bring salient prior of input images, bootstrapping the performance on a wide range of downstream tasks hence.

%% file: sections/3.method.tex
\section{Method}\label{sec:method}

In this section, we first give an overview of our proposed \method in \cref{sec:overview}.
Then, the two objectives in \method, \textit{i.e.}, reconstruction loss and predicting loss are introduced in \cref{sec:mim} and \cref{sec:loss_pred}, respectively.
Finally, in \cref{sec:mask}, the easy-to-hard mask generation manner is described.

\subsection{Overview}
\label{sec:overview}

Introduced in \cref{fig:motivation} and \cref{sec:intro}, conventional MVM pre-training solutions can be considered as training a student to solve \textit{given} problems,
while we argue that \textit{making the model stand in the shoes of a teacher}, producing challenging pretext task is crucial.
To achieve this, we introduce an auxiliary decoder to predict the reconstruction loss of each masked patch, and carefully design its objective.
\cref{fig:pipeline} gives an overview of our proposed \method, introduced next.

\method consists of a student ($f_{\theta_s}$, $d_{\phi_s}$, and $d_{\psi_s}$) and a teacher ($f_{\theta_t}$, $d_{\phi_t}$, and $d_{\psi_t}$) with the same network architecture.
$f_{\theta}(\cdot)$, $d_{\phi}(\cdot)$, and $d_{\psi}(\cdot)$ are encoder, visual reconstructor, and reconstruction loss predictor, parameterized by $\theta$, $\phi$, and $\psi$, respectively.
The subscript $t$ stands for teacher and $s$ stands for student.
To generate consistent predictions (especially for the reconstruction loss predictor), momentum update~\cite{he2020momentum} is applied to the teacher:
\begin{equation}
    \bm{\theta}_t \leftarrow m\bm{\theta}_t + (1 - m)\bm{\theta}_s,
\end{equation}
where $\bm{\theta}_t = (\theta_t, \phi_t, \psi_t)$, $\bm{\theta}_s = (\theta_s, \phi_s, \psi_s)$, and $m$ denotes the momentum coefficient.
We set $m=0.999$ in our experiments by default.

The input image or video can be represented as a 4D tensor $\mathbf{I} \in \mathbb{R}^{T\times H\times W \times 3}$, where $T$ is the temporal dimension, $H \times W$ is the spatial dimension, and 3 is the color channels.
Images are treated as being single-frame videos with $T=1$.
At each training iteration, input images or videos $\mathbf{I}$ are reshaped into a sequence of 2D patches $\mathbf{x} \in \mathbb{R}^{N\times(P^2C)}$, where $P$ and $t$ are the spatial patch size (\textit{i.e.}, 16) and the temporal patch size (\textit{i.e.}, 2), respectively, and $N=(T/t) \cdot (HW/P^2)$ is the number of patches hence.

Then, $\mathbf{x}$ is fed into the teacher to get the patch-wise predicted reconstruction loss matrix $\hat{\mathcal{L}}^t = d_{\psi_t}(f_{\theta_t}(\mathbf{x}))$.
Based on $\hat{\mathcal{L}}^t$ and the training status, a binary mask $\mathbf{M} \in \{0,1\}^{N}$ is generated under an easy-to-hard manner introduced later in \cref{sec:mask}.
The student is trained based on two objectives, \textit{i.e.}, reconstruction loss (\cref{sec:mim}) and predicting loss (\cref{sec:loss_pred})
\begin{equation}
    \mathcal{L} = \mathcal{L}_{\mathrm{rec}} + \mathcal{L}_{\mathrm{pred}},
\end{equation}
where these two objectives work in an alternating way, and reinforce each other to extract better representations by gradually urging the student to reconstruct hard patches within an image or video.

\subsection{Visual Reconstructor}
\label{sec:mim}
Masked visual modeling aims to train an autoencoder~\cite{hinton2006reducing} to reconstruct the masked portion according to pre-defined targets, \textit{e.g.}, raw RGB pixels~\cite{he2022masked, xie2022simmim, liu2022exploring, kong2022understanding, chen2022context, yi2022masked} and specific tokens~\cite{bao2021beit, zhou2021ibot, wei2022masked, baevski2022data2vec, wu2022extreme}.
\begin{equation}
\label{eq:mim}
    \mathcal{L}_{\mathrm{rec}} = \mathcal{M} \left(
    d_{\phi_s}(f_{\theta_s}(\mathbf{x} \odot \mathbf{M})), \mathcal{T}(\mathbf{x} \odot (1 - \mathbf{M})) 
    \right),
\end{equation}
where for conventional approaches, the binary mask $\mathbf{M} \in \{0,1\}^{N}$ is generated by a pre-defined manner.
$\odot$ means element-wise dot product, and thus $\mathbf{x} \odot \mathbf{M}$ represents unmasked (\textit{i.e.}, visible) patches and vice versa.
$\mathcal{T}(\cdot)$ is the transformation function, generating reconstructed targets.
$\mathcal{M}(\cdot,\cdot)$ represents the similarity measurement, \textit{e.g.}, normalized $\ell_2$-distance~\cite{he2022masked, tong2022videomae}, smooth $\ell_1$-distance~\cite{xie2022simmim}, knowledge distillation~\cite{zhou2021ibot, dong2022bootstrapped}, and cross-entropy~\cite{bao2021beit}.

\subsection{Hard Patches Mining with a Loss Predictor}
\label{sec:loss_pred}
In NLP, each word in a sentence is already highly semantic~\cite{he2022masked}.
Training a model to predict only a few missing words tends to be a challenging task in understanding languages~\cite{devlin2018bert, brown2020language, radford2018improving, radford2019language}.
In CV, on the contrary, visual signals usually have heavy spatial and temporal redundancy, and thus plenty of mask strategies are proposed to deal with this issue~\cite{he2022masked, tong2022videomae, bao2021beit, kakogeorgiou2022hide, shi2022adversarial, li2022semmae}.

Beyond generating a difficult problem by prior knowledge, we argue that \textit{the ability} to produce demanding scenarios is the key design.
Intuitively, we consider patches with \textit{high reconstruction loss} defined in \cref{eq:mim} as \textit{hard patches}, which implicitly indicates the most discriminative parts of the input verified in \cref{fig:discriminative}.
Therefore, if the model is equipped with the ability to predict the reconstruction loss for each patch, simply masking those hard patches becomes a more challenging pretext task.

To this end, we employ an extra loss predictor (\textit{i.e.}, $d_{\psi}$ in \cref{fig:pipeline}) to mine hard patches during training.
To train such a loss predictor, we design two variants: 1) absolute loss and 2) relative loss.
Next, the formulation of each variant will be introduced as follows.

\paragraph{Absolute loss.}
The simplest and the most straightforward way is to define the objective by computing the mean square error (MSE) over the predicted loss and the ground truth reconstruction loss:
\begin{equation}
\label{eq:mse}
    \mathcal{L}_{\mathrm{pred}} = \left(
    d_{\psi_s}(f_{\theta_s} (\mathbf{x} \odot \mathbf{M})) - \mathcal{L}_{\mathrm{rec}}
    \right)^2 \odot (1 - \mathbf{M}),
\end{equation}
where $d_{\psi_s}$ is the auxiliary decoder of the student parameterized by $\psi_s$, and $\mathcal{L}_{\mathrm{rec}}$ here is detached from gradient, being a ground-truth for loss prediction.
However, recall that our goal is to determine hard patches within an input, thus we need to learn the \textit{relative} relationship among patches.
Under such a setting, MSE is not the most suitable choice hence, since the scale of $\mathcal{L}_{\mathrm{rec}}$ decreases as training goes on, and thus the loss predictor may be overwhelmed by the scale and the exact value of $\mathcal{L}_{\mathrm{rec}}$.
For this purpose, we propose a binary cross-entropy-based relative loss as an alternative.
The detailed formulation of the proposed relative objective is described in the subsequent paragraph.

\paragraph{Relative loss.}
Intuitively, the patch-wise difficulty of the reconstruction task can be measured by $\texttt{argsort} (\mathcal{L}_{\mathrm{rec}})$.
Therefore, we aim to \textit{predict} $\texttt{argsort}(\mathcal{L}_{\mathrm{rec}})$ using a relative loss.
However, as the \texttt{argsort}$(\cdot)$ operation is non-differentiable, it is hard to directly minimize some custom distances between $\texttt{argsort}(d_{\psi_s}(f_{\theta_s} (\mathbf{x} \odot \mathbf{M})))$ and $\texttt{argsort}(\mathcal{L}_{\mathrm{rec}})$.
Therefore, we translate this problem into an equivalent one: \textit{dense relation comparison}.
Specifically, for each pair of patches $(i,j)$, where $i,j=1,2,\cdots,N$ and $i \neq j$, we can implicitly learn $\texttt{argsort}(\mathcal{L}_{\mathrm{rec}})$ by predicting the relative relation of $\mathcal{L}_{\mathrm{rec}}(i)$ and $\mathcal{L}_{\mathrm{rec}}(j)$, \textit{i.e.}, which one is larger.
The objective is defined as follows:
\begin{equation}
\label{eq:bce}
\begin{aligned}
    \mathcal{L}_{\mathrm{pred}} = 
    &-\sum_{i=1}^N \sum_{j=1 \atop j\neq i}^N \mathbbm{1}^{+}_{ij} \log \left( \sigma(\hat{\mathcal{L}}^s_i - \hat{\mathcal{L}}^s_j) \right) \\
    &-\sum_{i=1}^N \sum_{j=1 \atop j\neq i}^N \mathbbm{1}^{-}_{ij} \log \left( 1 - \sigma(\hat{\mathcal{L}}^s_i - \hat{\mathcal{L}}^s_j) \right),
\end{aligned}
\end{equation}
where $\hat{\mathcal{L}}^s = d_{\psi_s}(f_{\theta_s}(\mathbf{x} \odot \mathbf{M})) \in \mathbb{R}^N$ represents the predicted loss from the student, and $i, j=1,2,\dots,N$ are patch indexes.
$\sigma(\cdot)$ indicates \texttt{sigmoid} function, \textit{i.e.}, $\sigma(z) = e^z / (e^z + 1)$.
$\mathbbm{1}^{+}_{ij}$ and $\mathbbm{1}^{-}_{ij}$ are two indicators, representing the relative relationship of ground-truth reconstruction losses, \textit{i.e.}, $\mathcal{L}_{\mathrm{rec}}$, between patch $i$ and patch $j$
\begin{equation}
\label{eq:indicatorp}
\mathbbm{1}^{+}_{ij} = \left\{
\begin{aligned}
    &1, &&\mathcal{L}_{\mathrm{rec}}(i) > \mathcal{L}_{\mathrm{rec}}(j) \mathrm{\ and\ } \mathbf{M}_i=\mathbf{M}_j=0, \\
    &0, &&\mathrm{otherwise},
\end{aligned}
\right.
\end{equation}
\begin{equation}
\label{eq:indicatorn}
\mathbbm{1}^{-}_{ij} = \left\{
\begin{aligned}
    &1, &&\mathcal{L}_{\mathrm{rec}}(i) < \mathcal{L}_{\mathrm{rec}}(j) \mathrm{\ and\ } \mathbf{M}_i=\mathbf{M}_j=0, \\
    &0, &&\mathrm{otherwise},
\end{aligned}
\right.
\end{equation}
where $\mathbf{M}_i=\mathbf{M}_j=0$ means that both patch $i$ and $j$ are masked during training.


\subsection{Easy-to-Hard Masking}
\label{sec:mask}
With the reconstruction loss predictor, we are able to define a more challenging pretext task, \textit{i.e.}, mask those hard/discriminative parts of an input image.
Concretely, after obtaining the predicted reconstruction loss from the teacher network, \textit{i.e.}, $\hat{\mathcal{L}}^t = d_{\psi_t}(f_{\theta_t} (\mathbf{x}))$, we conduct $\texttt{argsort}(\cdot)$ operation over $\hat{\mathcal{L}}^t$ in a descending order to obtain the relative reconstruction difficulty.

However, in the early training stages, the learned feature representations may not be ready for reconstruction but are overwhelmed by the rich texture, which means large reconstruction loss may not be equivalent to discriminative.
To this end, we propose an easy-to-hard mask generation manner, providing some reasonable hints that guide the model to reconstruct masked hard patches step by step.

As illustrated in \cref{fig:pipeline}, for each training epoch $t$, $\alpha_t$ of the mask patches are generated by $\hat{\mathcal{L}}^t$, and the remaining $1 - \alpha_t$ are randomly selected.
$\alpha_t$ grows linearly from $\alpha_0$ to $\alpha_T$, which is defined as follows:
\begin{equation}
    \alpha_t = \alpha_0 + \frac{t}{T} (\alpha_T - \alpha_0),
\end{equation}
where $T$ is the total training epochs, and $\alpha_0,\alpha_T\in[0,1]$ are two tunable hyper-parameters.
We filter $\alpha_t\cdot\gamma N$ patches with the highest $\hat{\mathcal{L}}^t$ to be masked, and the remaining $(1-\alpha_t)\cdot\gamma N$ patches are randomly masked.
The proportion $\alpha_t$ gradually increases from $\alpha_0$ to $\alpha_T$ in a linear manner without further tuning for simplicity, contributing to an easy-to-hard training procedure.

%
The implementation of easy-to-hard masking is based on MAE~\cite{he2022masked}.
Specifically, at training epoch $t$, we want to generate a binary mask $\mathbf{M}$ with $\gamma N$ patches to be masked.
Under the easy-to-hard manner, there are $\alpha_t \gamma N$ patches masked by predicted loss $\hat{\mathcal{L}}^t$ and the remaining $(1 - \alpha_t) \gamma N$ are randomly selected.

%% file: sections/4.experiments.tex
\begin{table}[t]
    \centering
    \caption{
    Ablation study on different \textbf{reconstruction targets}.
    We study four different targets, including raw RGB pixels (MAE~\cite{he2022masked} baseline), and three knowledge distillation targets, \textit{i.e.}, features from the EMA (exponential moving average) model, DINO~\cite{caron2021emerging}, and CLIP~\cite{radford2021learning}.
    All cases are pre-trained 200 epochs on ImageNet~\cite{russakovsky2015imagenet} with ViT-B/16~\cite{dosovitskiy2020image}.
    For knowledge distillation cases, we minimize MSE between $\ell_2$ normalized features.
    ``Learn to mask'' indicates whether the mask is guided by predicted losses (if not, the mask is randomly generated). 
    %
    %
    }
    \label{tab:target}
    \vspace{-10pt}
    \setlength{\tabcolsep}{4pt}
    \begin{tabular}{llllll}
    \toprule
    \multirow{2}{*}{target} & \multirow{2}{*}{$\mathcal{L}_{\mathrm{pred}}$} & learn & \multirow{2}{*}{fine-tune} & \multirow{2}{*}{\gc{linear}} & \multirow{2}{*}{\gc{$k$-NN}} \\
    & & to mask & & & \\
    \midrule
    \multicolumn{6}{l}{\textit{Pixel Regression}} \\
    \midrule
    \multirow{3}{*}{RGB (MAE~\cite{he2022masked}} & - & - & 82.23 & \gc{50.80} & \gc{29.84} \\
    & \checkmark & - & 82.49 \up{0.26} & \gc{51.26} & \gc{31.98} \\
    & \cellcolor{Light}{\checkmark} & \cellcolor{Light}{\checkmark} & \cellcolor{Light}{\textbf{82.95 \up{0.72}}} & \cellcolor{Light}{\gc{\textbf{54.92}}} & \cellcolor{Light}{\gc{\textbf{36.09}}} \\
    \midrule
    \multicolumn{6}{l}{\textit{Feature Distillation}} \\
    \midrule
    \multirow{3}{*}{EMA features} & - & - & 82.99 & \gc{32.65} & \gc{20.69} \\
    & \checkmark & - & 83.13 \up{0.14} & \gc{52.06} & \gc{35.73} \\
    & \checkmark & \checkmark & \textbf{83.47 \up{0.48}} & \gc{\textbf{55.25}} & \gc{\textbf{35.94}} \\
    \midrule
    \multirow{3}{*}{DINO~\cite{caron2021emerging} features} & - & - & 83.46 & \gc{61.31} & \gc{41.53} \\
    & \checkmark & - & 83.58 \up{0.12} & \gc{63.25} & \gc{43.02} \\
    & \checkmark & \checkmark & \textbf{84.13 \up{0.67}} & \gc{\textbf{64.17}} & \gc{\textbf{47.25}} \\
    \midrule
    \multirow{3}{*}{CLIP~\cite{radford2021learning} features} & - & - & 83.20 & \gc{59.80} & \gc{42.51} \\
    & \checkmark & - & 83.31 \up{0.11} & \gc{60.62} & \gc{43.26} \\
    & \checkmark & \checkmark & \textbf{83.58 \up{0.38}} & \gc{\textbf{62.22}} & \gc{\textbf{45.08}} \\
    \bottomrule
    \end{tabular}
\end{table}

\begin{table}[t]
    \centering
    \caption{
    Ablation study on different \textbf{mask strategies}.
    We study the effect of different $\alpha_0$, $\alpha_T$, and $\gamma$.
    Large $\alpha_T$ indicates a more difficult pretext task, but the randomness of this strategy decreases.
    %
    }
    \label{tab:mask}
    \vspace{-10pt}
    \setlength{\tabcolsep}{3.3pt}
    \begin{tabular}{lccllll}
    \toprule
    case & difficulty & randomness & $\gamma$ & $\alpha_0$ & $\alpha_T$ & fine-tune \\
    \midrule
    random & easy & strong & 75 & 0 & 0 & 82.49 \\
    learn to mask & \multirow{2}{*}{$\Big\downarrow$} & \multirow{2}{*}{$\Big\downarrow$} & \cellcolor{Light}{75} & \cellcolor{Light}{0} & \cellcolor{Light}{0.5} & \cellcolor{Light}{\textbf{82.95 \up{0.46}}} \\
    learn to mask & & & 75 & 0 & 1 & 82.67 \up{0.18} \\
    learn to mask & hard & weak & 75 & 1 & 1 & 81.40 \down{1.09} \\
    \midrule
    random & easy & strong & 50 & 0 & 0 & 82.36 \\
    learn to mask & $\downarrow$ & $\downarrow$ & 50 & 0 & 0.5 & \textbf{82.56 \up{0.20}} \\
    learn to mask & hard & weak & 50 & 1 & 1 & 82.19 \down{0.17} \\
    \midrule
    random & easy & strong & 90 & 0 & 0 & 82.48 \\
    learn to mask & $\downarrow$ & $\downarrow$ & 90 & 0 & 0.5 & \textbf{82.66 \up{0.18}} \\
    learn to mask & hard & weak & 90 & 1 & 1 & 80.59 \down{1.89} \\
    \bottomrule
    \end{tabular}
\end{table}

\section{Experiments}\label{sec:exp}

We evaluate the effectiveness of our \method on both masked \textit{image} modeling and masked \textit{video} modeling benchmarks in \cref{sec:exp_mim} and \cref{sec:exp_mvm}, respectively.
Sufficient empirical evidence demonstrates that \method, as a unified framework, bootstraps masked visual modeling by making the model aware of the hard patches subsequently and producing challenge problems.

\subsection{Experiments on Masked Image Modeling}
\label{sec:exp_mim}

\noindent\textbf{Pre-training.}
Under masked \textit{image} modeling (MIM) benchmarks, we take ImageNet-1K~\cite{russakovsky2015imagenet}, which contains 1.3M images for 1K categories, as the pre-training dataset.
By default, we take ViT-B/16~\cite{dosovitskiy2020image} as the backbone and it is pre-trained 200 epochs (for ablations) followed by 100 epochs of end-to-end fine-tuning.
Most of the configurations are borrowed from MAE~\cite{he2022masked}.
The linear learning rate scaling rule~\cite{goyal2017accurate} is adopted: $lr = lr_{\mathrm{base}} \times \mathrm{batch\_size}\ /\ 256$.
Our implementation is based on MAE~\cite{he2022masked} and UM-MAE~\cite{li2022uniform}.

\paragraph{ImageNet classification.}
We evaluate our proposed \method by 1) end-to-end fine-tuning, 2) linear probing, and 3) $k$-NN.
We report Top-1 accuracy (\%) on the validation set.
End-to-end 100 epochs of fine-tuning is the main metric for evaluation.
For linear probing, we simply follow the setting of MoCo v3~\cite{chen2021empirical}, where we do not use mixup~\cite{zhang2017mixup}, cutmix~\cite{yun2019cutmix}, drop path~\cite{huang2016deep}, and color jitter.
The $k$-NN classification setting is borrowed from DINO~\cite{caron2021emerging}.
All images are first resized to 256$\times$256 and then center-cropped to 224$\times$224.
We report the best result among $k=10,20,100,200$.

\paragraph{COCO object detection and instance segmentation.}
We take Mask R-CNN~\cite{he2017mask} with FPN~\cite{lin2017feature} as the object detector.
Following~\cite{he2022masked} and~\cite{li2022uniform}, to obtain pyramid feature maps for matching the requirements of FPN~\cite{lin2017feature}, whose feature maps are all with a stride of 16, we equally divide the backbone into 4 subsets, each consisting of a last global-window block and several local-window blocks otherwise, and then apply convolutions to get the intermediate feature maps at different scales (stride 4, 8, 16, or 32), which is the same as ResNet~\cite{he2016deep}.
We perform end-to-end fine-tuning on COCO~\cite{lin2014microsoft} for 1$\times$ schedule, \textit{i.e.}, 12 epochs, for ablations (\textit{i.e.}, Tab. \textcolor{red}{6}) with 1024$\times$1024 resolution.
We simply follow the configuration of ViTDet~\cite{li2021benchmarking} in detectron2~\cite{wu2019detectron2}.
Experiments are conducted on 8 Telsa V100 GPUs with a batch size of 16.

\paragraph{ADE20k semantic segmentation.}
We take UperNet~\cite{xiao2018unified} as the segmentor, and perform end-to-end fine-tuning on ADE20k~\cite{zhou2017scene} for 80k iterations for ablations (\textit{i.e.}, \cref{tab:downstream}) and 160k iterations when comparing with previous methods (\textit{i.e.}, \cref{tab:sota_seg}) with 512$\times$512 resolution.
We take mIoU~\cite{everingham2015pascal} as the evaluation metric.
Our implementation is based on mmsegmentation~\cite{mmseg2020}.
The AdamW~\cite{loshchilov2017decoupled} optimizer is adopted with an initial learning rate of 1e-4 and a weight decay of 0.05 with ViT-B.
For ViT-L, the learning rate is 2e-5.
We apply a polynomial learning rate schedule with the first warmup of 1500 iterations following common practice~\cite{li2022uniform, mmseg2020, bao2021beit}.
Experiments are conducted on 8 Telsa V100 GPUs.

\paragraph{Effective training epochs.}
Following iBOT~\cite{zhou2021ibot}, we take the effective training epochs as the metric of the training schedule, due to extra computation costs brought by multi-crop~\cite{caron2020unsupervised} augmentation,
which is a widely used technique for contrastive methods.
Specifically, the effective training epochs are defined as the actual pre-training
epochs multiplied with a scaling factor $r$.
For instance, DINO~\cite{caron2021emerging} is trained with 2 global 224$\times$224 crops and 10 local 96$\times$96 crops, and thus $r=2 + (96/224)^2 \times 10 \approx 4$.
More details and examples can be found in~\cite{zhou2021ibot}.

\subsubsection{Ablation Study}
\label{sec:ablation}

We study different reconstruction targets, mask strategies, predicting loss formulations, and downstream tasks in this section.
By default, ViT-B/16~\cite{dosovitskiy2020image} is used as the backbone with 200 epochs pre-training and 100 epochs fine-tuning on ImageNet-1K~\cite{russakovsky2015imagenet}.
We \colorbox{Light}{highlight} our default settings.
%

\paragraph{Reconstruction targets.}
We study the effectiveness of different reconstruction targets in \cref{tab:target}, including regressing raw RGB pixels used in MAE~\cite{he2022masked}, and distilling from various teacher models, \textit{i.e.}, the EMA (exponential moving average) teacher used in BootMAE~\cite{dong2022bootstrapped}, and pre-trained teachers obtained from DINO~\cite{caron2021emerging} and CLIP~\cite{radford2021learning}.
All these teacher models share the same architecture, \textit{i.e.}, ViT-B/16~\cite{dosovitskiy2020image}.

It has been substantiated that directly regressing RGB values of pixels is a simple yet efficient way in MIM pre-training~\cite{he2022masked}.
However, due to the existence of high-frequency noise in some cases, patches with higher frequency tend to have larger reconstruction loss, and thus \textit{hard patches may not be highly semantic} under this setting, which is quite the opposite from our motivation: learn to mine \textit{discriminative} parts of an image instead of high-frequency parts.
To this end, we further take features from a teacher model to be the learning target (\textit{e.g.}, DINO~\cite{caron2021emerging} and CLIP~\cite{radford2021learning}), to verify the effectiveness of our proposed \method.
%

Note that the objective differs when using different reconstruction targets.
Specifically, an MSE loss is adopted for RGB regression following MAE~\cite{he2022masked}, while for knowledge distillation cases, we first apply $\ell_2$ normalization to the features output from the teacher and the student, and then minimize their MSE distances.
This can be also implemented by maximizing their cosine similarities.

\begin{table}[t]
    \centering
    \caption{
    Ablation study on different \textbf{mask strategies}.
    We study the effectiveness of the $\texttt{argmax}(\cdot)$ performed on predicted reconstruction loss $\hat{\mathcal{L}}^t$ and the ``easy-to-hard'' manner.
    Note that $\texttt{argmin}(\cdot)$ means that we mask those easy patches.
    %
    }
    \label{tab:mask_manner}
    \vspace{-10pt}
    \begin{tabular}{llllll}
    \toprule
    case & operation & $\gamma$ & $\alpha_0$ & $\alpha_T$ & fine-tune \\
    \midrule
    random & - & 75 & 0 & 0 & 82.49 \\
    \rowcolor{Light}
    learn to mask & \texttt{argmax}$(\cdot)$ & 75 & 0 & 0.5 & \textbf{82.95 \up{0.46}} \\
    learn to mask & \texttt{argmin}$(\cdot)$ & 75 & 0 & 0.5 & 82.36 \down{0.13} \\
    \midrule
    \midrule
    case & manner & $\gamma$ & $\alpha_0$ & $\alpha_T$ & fine-tune \\
    \midrule
    random & - & 75 & 0 & 0 & 82.49 \\
    \rowcolor{Light}
    learn to mask & easy-to-hard & 75 & 0 & 0.5 & \textbf{82.95 \up{0.46}} \\
    learn to mask & hard-to-easy & 75 & 0.5 & 0 & 81.71 \down{0.78} \\
    \bottomrule
    \end{tabular}
\end{table}

\begin{table}[t]
    \centering
    \caption{
    Ablations on \textbf{predicting loss formulations}.
    We study the absolute loss introduced in \cref{eq:mse} and the relative loss described in \cref{eq:bce}.
    %
    }
    \vspace{-10pt}
    \setlength{\tabcolsep}{13pt}
    \begin{tabular}{llll}
    \toprule
    case & fine-tune & \gc{linear} & \gc{$k$-NN} \\
    \midrule
    none (MAE~\cite{he2022masked}) & 82.23 & \gc{51.26} & \gc{31.98} \\
    absolute MSE & 82.77 \up{0.54} & \gc{51.85} & \gc{34.47} \\
    \rowcolor{Light}
    relative BCE & \textbf{82.95 \up{0.72}} & \textbf{\gc{54.92}} & \textbf{\gc{36.09}} \\
    \bottomrule
    \end{tabular}
\end{table}

As illustrated in \cref{tab:target}, our \method is able to bootstrap the performances under various learning targets.
Taking the pixel regression case as an instance, equipped with the predicting loss and the easy-to-hard mask generation manner, the fine-tuning Top-1 accuracy achieves 82.95\%, outperforming MAE~\cite{he2022masked} by +0.72\%.
Notably, \textit{only} applying an auxiliary decoder to predict reconstruction loss for each patch brings an improvement of +0.26\% fine-tuning accuracy, achieving 82.49\%, verifying that \textit{the ability to mine hard patches} brings better extracted feature representations.
Then, fully taking advantage of this capability, \textit{i.e.}, generate challenging masks, can further bootstrap the performances,
which appears \textit{consistently across different learning targets}.

\paragraph{Mask strategies.}
To verify that harder tasks do bring better performance, we study various mask strategies in \cref{tab:mask}, including random masking and our proposed learnable masking.
With different $\alpha_0$ and $\alpha_T$, we can construct different strategies.
For instance, $\alpha_0=\alpha_T=0$ indicates that predicted reconstruction losses $\hat{\mathcal{L}}^t$ will not participate in mask generation (\textit{i.e.}, a fully random manner),
$\alpha_0=\alpha_T=1$, however, means that $\gamma N$ patches with the highest $\hat{\mathcal{L}}^t$ values are kept masked (see \cref{fig:discriminative}).

From \cref{tab:mask}, we find that the increase in the difficulty of the pretext task does not consistently lead to better performance.
\textit{Retaining a certain degree of randomness} is beneficial for satisfactory results.
Specifically, $\alpha_0=0$ and $\alpha_T=0.5$ achieves the best results under different mask ratio $\gamma$, which is a more difficult case over $\alpha=\alpha_T=0$ (\textit{i.e.}, random masking), and with stronger randomness against $\alpha_0=\alpha_T=1$.
These conclusions are quite intuitive.
Directly masking those patches with the highest $\hat{\mathcal{L}}^t$ brings the hardest problem, where discriminative parts of an image are almost masked.
That means visible patches are nearly all background (see \cref{fig:discriminative}).
\textit{Forcing the model to reconstruct the forehead based on only these backgrounds without any hints makes no sense},
whose performance drops consistently with different values of $\gamma$.
Therefore, a certain level of randomness is necessary.
%

We further investigate the effectiveness of producing \textit{hard} pretext task for MIM pre-training in \cref{tab:mask_manner}.
Note that performing $\texttt{argmin}(\cdot)$ operation over predicted reconstruction loss $\hat{\mathcal{L}}^t$ means we have generated a task even easier than the random baseline.
$\alpha_0 < \alpha_T$ indicates an easy-to-hard mask generation introduced in \cref{sec:mask}, while $\alpha_0 > \alpha_T$ means the opposite, \textit{i.e.}, a hard-to-easy manner, which is also studied in \cref{tab:mask_manner}.
All results verify the necessity of a hard pretext task and the easy-to-hard manner.
Both \texttt{argmin}$(\cdot)$ operation and the hard-to-easy mask generation manner lead to performance degradation over random masking.

\paragraph{Predicting loss formulations.}
We study different designs of predicting loss in the following table, including absolute loss based on MSE introduced in \cref{eq:mse} and relative loss based on BCE defined in \cref{eq:bce}.
As expected, BCE is a better choice for mining \textit{relative relationship} between patches, instead of absolute values of reconstruction losses as MSE does, outperforming absolute MSE by +0.18\%.

\begin{table}[t]
    \centering
    \caption{
    Ablation study on different \textbf{decoder designs}.
    The speedup is evaluated under 8 Telsa V100 GPUs with 32 images with resolution 224$\times$224 per GPU.
    %
    }
    \label{tab:decoder}
    \vspace{-10pt}
    \setlength{\tabcolsep}{12pt}
    \begin{tabular}{ccccc}
    \toprule
    \# blocks & speedup & fine-tune & \gc{linear} & \gc{$k$-NN} \\
    \midrule
    1 & 1.94$\times$ & 82.67 & \gc{39.83} & \gc{16.83} \\
    2 & 1.68$\times$ & 82.50 & \gc{46.74} & \gc{22.63} \\
    4 & 1.37$\times$ & 82.75 & \gc{53.95} & \gc{33.60} \\
    \rowcolor{Light}
    8 & 1.00$\times$ & \textbf{82.95} & \gc{54.92} & \gc{36.09} \\
    12 & 0.76$\times$ & 82.84 & \gc{54.83} & \gc{35.93} \\
    \midrule
    \midrule
    \# dim & speedup & fine-tune & \gc{linear} & \gc{$k$-NN} \\
    \midrule
    128 & 1.31$\times$ & 82.74 & \gc{42.51} & \gc{17.67} \\
    256 & 1.18$\times$ & 82.80 & \gc{52.39} & \gc{29.46} \\
    \rowcolor{Light}
    512 & 1.00$\times$ & \textbf{82.95} & \gc{54.92} & \gc{36.09} \\
    1024 & 0.61$\times$ & 82.81 & \gc{54.01} & \gc{36.54} \\
    \bottomrule
    \end{tabular}
\end{table}

\begin{wraptable}{r}{0.35\linewidth}
    \centering\footnotesize
    \vspace{-10pt}
    \addtolength\leftskip{-8pt}
    \begin{tabular}{ll}
    method & fine-tune \\
    \shline
    BEiT~\cite{bao2021beit} & 80.9 \\
    \ + \method & \textbf{81.5 \up{0.6}} \\
    iBOT~\cite{zhou2021ibot} & 82.9 \\
    \ + \method & \textbf{83.4 \up{0.5}} \\
    \end{tabular}
    \vspace{-10pt}
\end{wraptable}

\paragraph{Baselines.}
We study the effectiveness of \method over BEiT~\cite{bao2021beit} and iBOT~\cite{zhou2021ibot} in the right table.
We perform 200 and 50 epochs pre-training for BEiT~\cite{bao2021beit} and iBOT~\cite{zhou2021ibot}, respectively.
Note that iBOT~\cite{zhou2021ibot} utilizes 2 global crops ($224^2$) and 10 local crops ($96^2$).
Therefore, the effective pre-training epoch of iBOT-based experiments is $50\times(2+\frac{10\times96^2}{224^2}) \approx 200$. 
From the table, we can tell that \method brings consistent and significant improvements.

\begin{table}[t]
    \centering
    \caption{
    Ablations on \textbf{downstream tasks}.
    We take RGB and CLIP~\cite{radford2021learning} features as the learning target, representing \textit{pixel regression} and \textit{knowledge distillation} cases.
    All cases are first pre-trained 200 epochs on ImageNet-1K~\cite{russakovsky2015imagenet} with ViB-B/16~\cite{dosovitskiy2020image} followed by fine-tuning.
    ``Learn to mask'' indicates whether the mask is guided by predicted losses (if not, the mask is randomly generated). 
    %
    }
    \label{tab:downstream}
    \vspace{-10pt}
    \setlength{\tabcolsep}{2.8pt}
    \begin{tabular}{llllll}
    \toprule
    \multirow{2}{*}{target} &
    \multirow{2}{*}{$\mathcal{L}_{\mathrm{pred}}$} &
    learn &
    \multicolumn{2}{l}{COCO} & 
    \multicolumn{1}{l}{ADE20k} \\
    \cline{4-6}
    & & to mask & AP$_{\text{box}}$ & AP$_{\text{mask}}$ & mIoU \\
    \midrule
    \multirow{3}{*}{RGB} & 
    - & -  & 40.45 & 37.01 & 40.49 \\
    & \checkmark & - & 40.98 \up{0.53} & 37.34 \up{0.33} & 41.45 \up{0.96} \\
    & \cellcolor{Light}{\checkmark} & \cellcolor{Light}{\checkmark} & \cellcolor{Light}{\textbf{42.03 \up{1.58}}} & \cellcolor{Light}{\textbf{38.15 \up{1.14}}} & \cellcolor{Light}{\textbf{42.09 \up{1.60}}} \\
    \midrule
    \multirow{3}{*}{CLIP~\cite{radford2021learning}} & - & -  & 46.21 & 41.55 & 46.59 \\
    & \checkmark & - & 46.43 \up{0.22} & 41.80 \up{0.25} & 46.97 \up{0.38} \\
    & \checkmark & \checkmark & \textbf{46.57 \up{0.36}} & \textbf{41.96 \up{0.41}} & \textbf{47.35 \up{0.76}} \\
    \bottomrule
    \end{tabular}
\end{table}

\paragraph{Decoder designs.}
Our decoder is a stack of Transformer blocks~\cite{vaswani2017attention} with a fixed width following~\cite{he2022masked}.
We study its depth and width in \cref{tab:decoder}.
8 blocks with 512-d features is the best choice, which is exactly the same with MAE~\cite{he2022masked}.

\paragraph{Downstream tasks.}
We evaluate transfer learning performance using the pre-trained models in \cref{tab:target}, including COCO~\cite{lin2014microsoft} object detection and instance segmentation, and ADE20k~\cite{zhou2017scene} semantic segmentation.

As illustrated in \cref{tab:downstream}, equipped with our proposed \method, it outperforms +1.58 AP$_{\text{box}}$ and +1.14 AP$_{\text{mask}}$ on COCO~\cite{lin2014microsoft}, and +1.60 mIoU on ADE20k~\cite{zhou2017scene}, over MAE~\cite{he2022masked} baseline, \textit{i.e.}, taking raw RGB pixel as the learning target.
When using CLIP~\cite{radford2021learning} features as the learning target, it outperforms +0.36 AP$_{\text{box}}$ and +0.41 AP$_{\text{mask}}$ on COCO~\cite{lin2014microsoft}, and +0.76 mIoU on ADE20k~\cite{zhou2017scene} over baseline, respectively.

Notably, \textit{only} taking the predicting loss $\mathcal{L}_{\mathrm{pred}}$ as the extra objective manages to boost the performance across downstream tasks, verifying the effectiveness of making the model be the teacher, instead of only a student.
These observations are consistent across different learning targets.

\begin{table}[t]
    \centering
    \caption{
    Comparison with state-of-the-art alternatives on \textbf{ImageNet-1K}.
    All methods are evaluated by fine-tuning.
    The resolution of images is 224$\times$224 for both pre-training and fine-tuning.
    $\dag$ means our implementation.
    $\ddag$ means the result is borrowed from~\cite{he2022masked}.
    %
    }
    \label{tab:sota}
    \vspace{-10pt}
    \setlength{\tabcolsep}{10pt}
    \begin{tabular}{ll cll}
    \toprule
    method & venue & eff. ep. & ViT-B & ViT-L \\
    \midrule
    \gc{scratch} & & \gc{-} & \gc{80.9$^{\dag}$} & \gc{82.6$^{\ddag}$} \\
    \midrule
    \multicolumn{5}{l}{\textit{Contrastive Learning}} \\
    MoCo v3$^{\ddag}$~\cite{chen2021empirical} & \pub{ICCV'21} & 600 & 83.2 & 84.1 \\
    DINO$^{\ddag}$~\cite{caron2021emerging} & \pub{ICCV'21} & 1600 & 83.6 & - \\
    \midrule
    \multicolumn{5}{l}{\textit{MIM with Pixel Regression}} \\
    MAE~\cite{he2022masked} & \pub{CVPR'22} & 200 & 82.2$^{\dag}$ & 83.3$^{\ddag}$ \\
    \rowcolor{Light}
    \method & \pub{Ours} & 200 & \textbf{83.0} & \textbf{84.5} \\
    MAE$^{\ddag}$~\cite{he2022masked} & \pub{CVPR'22} & 1600 & 83.6 & 85.1 \\
    SimMIM~\cite{xie2022simmim} & \pub{CVPR'22} & 800 & 83.8 & - \\
    \rowcolor{Light}
    \method & \pub{Ours} & 800 & \textbf{84.2} & \textbf{85.8} \\
    \midrule
    \multicolumn{5}{l}{\textit{MIM with Feature Distillation}} \\
    BEiT$^{\ddag}$~\cite{bao2021beit} & \pub{ICLR'22} & 800& 83.2 & 85.2 \\
    iBOT~\cite{zhou2021ibot} & \pub{ICLR'22} & 1600 & 84.0 & - \\
    BootMAE~\cite{dong2022bootstrapped} & \pub{ECCV'22} & 800 & 84.2 & 85.9 \\
    \bottomrule
    \end{tabular}
\end{table}

\begin{table}[t]
    \centering
    \caption{
    Comparison with state-of-the-art alternatives on \textbf{ADE20k semantic segmentation} using UperNet.
    We take mIoU as the metric.
    $\ddag$ means the result is borrowed from~\cite{he2022masked}.
    }
    \label{tab:sota_seg}
    \vspace{-10pt}
    \setlength{\tabcolsep}{10pt}
    \begin{tabular}{llcc}
    \toprule
    method & venue & ViT-B & ViT-L \\
    \midrule
    \gc{supervised$^{\ddag}$} & & \gc{47.4} & \gc{49.9} \\
    MoCo v3$^{\ddag}$~\cite{chen2021empirical} & \pub{ICCV'21} & 47.3 & 49.1 \\
    BEiT$^{\ddag}$~\cite{bao2021beit} & \pub{ICLR'22} & 47.1 & 53.3 \\
    MAE$^{\ddag}$~\cite{he2022masked} & \pub{CVPR'22} & 48.1 & 53.6 \\
    SemMAE~\cite{li2022semmae} & \pub{NeurIPS'22} & 46.3 & - \\
    \rowcolor{Light}
    \method & \pub{Ours} & \textbf{48.5} & \textbf{54.6} \\
    \bottomrule
    \end{tabular}
\end{table}

\subsubsection{Comparison with Previous Alternatives}
\label{sec:sota}

We compare our proposed \method with the supervised baseline and a wide range of self-supervised alternatives using fine-tuning accuracy in \cref{tab:sota},
where selected methods can be summarized into three mainstream: (1) contrastive learning methods~\cite{chen2021empirical, caron2021emerging}, (2) MIM with pixel regression methods~\cite{he2022masked, xie2022simmim}, and (3) MIM with feature distillation methods~\cite{zhou2021ibot, dong2022bootstrapped, bao2021beit}.
Effective pre-training epoch is used for fair comparison following~\cite{zhou2021ibot}.
%
%
All methods are evaluated under the same input size \textit{i.e.}, 224$\times$224.
We take raw RGB as the learning target following~\cite{he2022masked, xie2022simmim}.

Notably, with only 200 epochs pre-training, our \method achieves 83.0\% and 84.5\% Top-1 accuracy with ViT-B and ViT-L backbone, respectively, 
surpassing MAE~\cite{he2022masked} by +0.8\% and +1.2\%, and the supervised baseline by +2.1\% and +1.9\%, respectively.
With a longer training schedule, \textit{i.e.}, 800 epochs, \method achieves 84.2\% and 85.8\% Top-1 accuracy with ViT-B and ViT-L backbone, outperforming MAE~\cite{he2022masked} by +0.6\% and +0.7\%, respectively.
Strikingly, \method reaches comparable results with \textit{feature distillation} alternative BootMAE~\cite{dong2022bootstrapped}.
From \cref{tab:target}, taking EMA features as the learning target for \method, which is the same as BootMAE~\cite{dong2022bootstrapped}, can further improve the performance by $\sim$ 0.5\%.


\begin{table}[t]
\centering
\caption{
In-depth analysis of the relationship between \textbf{hard to reconstruct} and \textbf{discrimination for classification}.
We train three ViT-B/16~\cite{dosovitskiy2020image} models \textit{from scratch} on ImageNet-1K~\cite{russakovsky2015imagenet} for 100 epochs under image-level supervision \textit{with only 50\% patches as input}.
``Case'' indicates which type of patches are visible, respectively.
}
\label{tab:in-depth}
\vspace{-10pt}
\setlength{\tabcolsep}{15pt}
\begin{tabular}{lll}
    \toprule
    case & portion & accuracy \\
    \midrule
    random & 50\% & 79.1 \\
    low $\hat{\mathcal{L}}^t$ & 50\% & 78.7 \down{0.4} \\
    high $\hat{\mathcal{L}}^t$ & 50\% & \textbf{79.8 \up{0.7}} \\
    \gc{all} & \gc{100\%} & \gc{80.9} \\
    \bottomrule
\end{tabular}
\end{table}

\begin{figure}[t]
    \centering
    \includegraphics[width=1\linewidth]{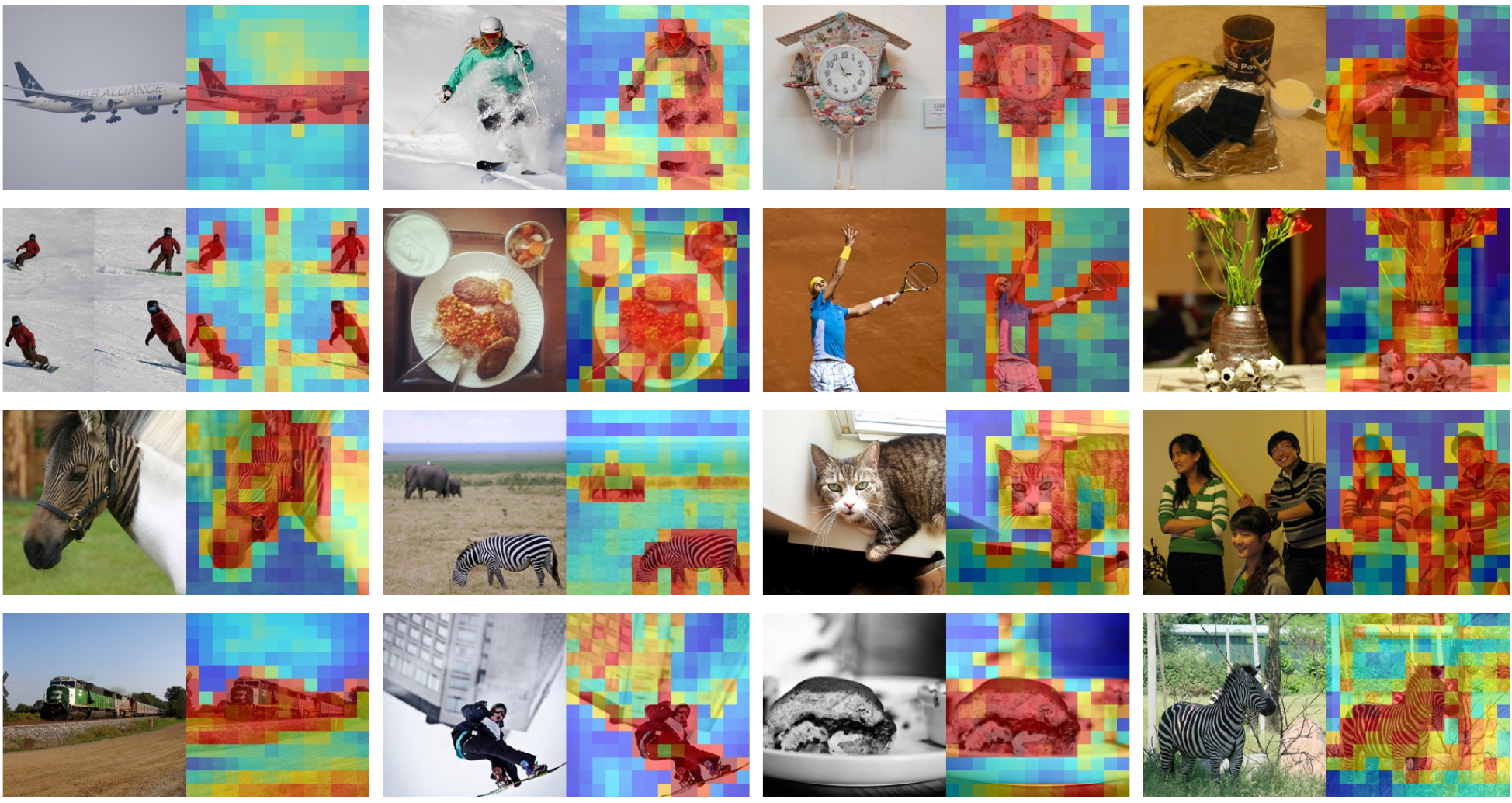}
    \vspace{-20pt}
    \caption{
    Visualization on \textbf{COCO} \textit{validation} set.
    For each tuple, we show the \textit{image} (left) and \textit{predicted} reconstruction losses (right).
    Red means higher losses and blue indicates the opposite.
    }
    \label{fig:coco_half}
\end{figure}

\paragraph{Semantic Segmentation.}
We experiment on ADE20k~\cite{zhou2017scene} using UperNet~\cite{xiao2018unified} for 160k iterations  in \cref{tab:sota_seg}.
%
%
From the table, we can tell that our \method significantly improves performance over supervised pre-training by +1.1 mIoU (48.5 \textit{v.s.} 47.4) with ViT-B and +4.7 mIoU (54.6 \textit{v.s.} 49.9) with ViT-L, respectively.
More importantly, our \method outperforms self-supervised alternatives under all settings.
For example, with ViT-L, \method surpasses MAE~\cite{he2022masked} by +1.0 (54.6 \textit{v.s.} 53.6) mIoU.

\begin{table}[t]
    \centering
    \caption{
    Ablation study on different \textbf{mask strategies}.
    We study the effectiveness of the $\texttt{argmax}(\cdot)$ operation on the predicted loss $\hat{\mathcal{L}}^t$ and the mask ratio.
    $\texttt{argmin}(\cdot)$ means we mask those easy patches.
    }
    \label{tab:mask_video}
    \vspace{-10pt}
    \begin{tabular}{lllll}
    \toprule
    case & operation & $\gamma$ & top-1 & top-5 \\
    \midrule
    random & - & 90 & 66.86 & 90.66 \\
    \rowcolor{Light}
    learn to mask & \texttt{argmax}$(\cdot)$ & 90 & \textbf{68.02 \up{1.16}} & \textbf{91.13 \up{0.47}} \\
    learn to mask & \texttt{argmin}$(\cdot)$ & 90 & 66.76 \down{0.10} & 90.17 \down{0.49} \\
    \midrule
    random & - & 75 & 65.15 & 89.77 \\
    learn to mask& \texttt{argmax}$(\cdot)$ & 75 & \textbf{66.32 \up{1.20}} & \textbf{90.40 \up{0.63}} \\
    learn to mask & \texttt{argmin}$(\cdot)$ & 75 & 65.17 \up{0.02} & 89.22 \down{0.55} \\
    \bottomrule
    \end{tabular}
\end{table}

\begin{table}[t]
    \centering
    \caption{
    Ablation study on \textbf{easy-to-hard masking}.
    Different schedule of $\alpha$ represents different cases.
    When $\alpha$ \textit{increases}, it follows an easy-to-hard schedule.
    Otherwise, it is a hard-to-easy schedule.
    }
    \label{tab:easy_to_hard_video}
    \vspace{-10pt}
    \setlength{\tabcolsep}{4pt}
    \begin{tabular}{llllll}
    \toprule
    case & manner & $\alpha_0$ & $\alpha_T$ & top-1 & top-5 \\
    \midrule
    random & - & 0 & 0 & 66.86 & 90.66 \\
    \rowcolor{Light}
    learn to mask & easy-to-hard & 0 & 0.5 & \textbf{68.02 \up{1.16}} & \textbf{91.13 \up{0.47}} \\
    learn to mask & easy-to-hard & 0 & 0.75 & 66.87 \up{0.01} & 90.75 \up{0.09} \\
    learn to mask & hard-to-easy & 0.5 & 0 & 66.59 \down{0.27} & 90.05 \down{0.61} \\
    \bottomrule
    \end{tabular}
\end{table}

\begin{table}[t]
    \centering
    \caption{
    Ablation study on the \textbf{predicting loss}.
    ``Learn to mask'' indicates the mask is generated based on predicted losses introduced in \cref{sec:mask}. 
    }
    \label{tab:loss_video}
    \vspace{-10pt}
    \begin{tabular}{llll}
    \toprule
    $\mathcal{L}_{\mathrm{pred}}$ & mask & top-1 & top-5 \\
    \midrule
    none & random & 66.86 & 90.66 \\
    absolute MSE & random & 67.22 \up{0.36} & 90.79 \up{0.13} \\
    absolute MSE & learn to mask & 67.74 \up{0.88} & 91.03 \up{0.37} \\
    relative BCE & random & 67.60 \up{0.74} & 90.81 \up{0.15} \\
    \rowcolor{Light}
    relative BCE & learn to mask & \textbf{68.02 \up{1.16}} & \textbf{91.13 \up{0.47}} \\
    \bottomrule
    \end{tabular}
\end{table}

\subsubsection{In-Depth Analysis}

\noindent\textbf{Hard to reconstruct \textit{v.s.} discrimination for classification.}
We present a pilot experiment in \cref{tab:in-depth} to explore the relationship between \textit{hard to reconstruct} and \textit{discrimination for classification}.
Three ViT-B/16~\cite{dosovitskiy2020image} models with only 50\% input tokens are trained from scratch on ImageNet-1K~\cite{russakovsky2015imagenet} for 100 epochs under image-level supervision.
The selection of visible patches differs.
We load HPM pre-trained with 200 epochs for computing the predicted loss $\hat{\mathcal{L}}^t$ and then choose visible tokens based on $\hat{\mathcal{L}}^t$.
Empirically, patches with higher $\hat{\mathcal{L}}^t$ contribute more to classification, indicating that these patches tend to be more discriminative.
We hope this will inspire future work.

\paragraph{Visualization of predicted losses.}
We provide qualitative results on COCO~\cite{lin2014microsoft} \textit{validation} set in \cref{fig:coco_half} using \method pre-trained with 200 epochs on ImageNet-1K~\cite{russakovsky2015imagenet}.
The model has \textit{never} seen this dataset.
Patches with higher \textit{predicted} reconstruction loss usually are more discriminative.

\subsection{Experiments on Masked Video Modeling}
\label{sec:exp_mvm}

\noindent\textbf{Pre-training.}
Under masked \textit{video} modeling benchmarks, we perform pre-training on both Kinetics-400 (K400)~\cite{kay2017kinetics} and Something-Something v2 (SSv2)~\cite{goyal2017something}, respectively.
K400~\cite{kay2017kinetics} contains around 240k training videos and 20k validation videos of 10s from 400 classes.
SSv2~\cite{goyal2017something} is another large-scale video dataset, having around 169k videos for training and 20k videos for validation and containing 174 motion-centric action classes.

By default, we use ViT-B/16~\cite{dosovitskiy2020image} with the join space-time attention~\cite{arnab2021vivit, liu2022video} pre-trained with 400 epochs on Something-Something v2~\cite{goyal2017something} for ablations.
The temporal patch size is 2~\cite{tong2022videomae, feichtenhofer2022masked, wei2022masked, arnab2021vivit} and a spatial patch size of 16$\times$16~\cite{dosovitskiy2020image, he2022masked}, denoted as 2$\times$16$\times$16.
For a 16$\times$224$\times$224 input, this patch size produces 8$\times$14$\times$14 tokens.
The pre-training configuration mostly follows VideoMAE~\cite{tong2022videomae}.

\paragraph{Evaluation.}
We evaluate the effectiveness of \method by fine-tuning the pre-trained backbone.
Top-1 and Top-5 accuracies (\%) are reported.
Following VideoMAE~\cite{tong2022videomae}, \method is evaluated with 5 clips $\times$ 3 crops on K400~\cite{kay2017kinetics}, and 2 clips $\times$ 3 crops on SSv2~\cite{goyal2017something}, respectively.
%

\subsubsection{Ablation Study}

To further investigate whether there are necessary modifications when adapting \method from the image domain to the video domain, we conduct sufficient experiments to ablate the best setting of \method under masked video modeling benchmarks.
Specifically, we study different mask strategies, the easy-to-hard masking manner, and predicting loss formulations.
For efficient evaluation, ViT-B/16~\cite{dosovitskiy2020image} is used as the backbone with 400 epochs pre-training and 50 epochs fine-tuning on Something-Something v2~\cite{goyal2017something}.
We \colorbox{Light}{highlight} our default settings.

Interestingly, empirical evidence suggests that \textit{almost no} modification is necessary, demonstrating that \method can be a \textit{unified} paradigm that bootstraps masked visual modeling with minimal domain knowledge and few inductive biases.

\paragraph{Mask strategies.}
We study the effectiveness of different mask strategies in \cref{tab:mask_video}, including, the $\texttt{argmax}(\cdot)$ operation, \textit{i.e.}, whether more difficult problems contribute to better representations, and the mask ratio.

First, we study the efficacy of the $\texttt{argmax}(\cdot)$ operation.
Specifically, this operation stands for we mask those patches with \textit{high} values of $\hat{\mathcal{L}}^t$, \textit{i.e.}, \textit{hard} patches are masked.
On the opposite, the $\texttt{argmin}(\cdot)$ indicates \textit{easy} problems (even easier than the random masking baseline).
With different mask ratios, masking hard patches, \textit{i.e.}, applying the $\texttt{argmax}(\cdot)$ operation, consistently outperforms other counterparts, demonstrating that a difficult pretext task is necessary for MVM.
Furthermore, a pretext task easier than the random masking baseline, \textit{i.e.}, by applying the $\texttt{argmin}(\cdot)$ operation, sometimes leads to performance degradation.

Next, the effectiveness of different mask ratios is explored.
The conclusion is similar to prior works~\cite{feichtenhofer2022masked, tong2022videomae}, where a large mask ratio becomes crucial.

\paragraph{Easy-to-hard masking.}
We study the effectiveness of easy-to-hard masking in \cref{tab:easy_to_hard_video}.
Different values of $\alpha_0$ and $\alpha_T$ control the masking manner.
Specifically, $\alpha_0 < \alpha_T$ means an easy-to-hard manner, while $\alpha_0 > \alpha_T$ indicates a hard-to-easy manner.
A larger $\alpha_T$ means a more difficult task.
From \cref{tab:easy_to_hard_video} we can conclude that the easy-to-hard manner is effective and the pretext task should not be too hard.

\paragraph{Predicting loss.}
We ablate the formulation of the predicting loss and study the effectiveness of the ability to mine hard patches (``learn to mask'') in \cref{tab:loss_video}.
The absolute MSE loss and the relative BCE loss are introduced in \cref{eq:mse} and \cref{eq:bce}, respectively.
As demonstrated in the table, BCE is a better choice to mine \textit{relative relationships} between patches.
Furthermore, \textit{solely} applying the auxiliary objective (either MSE or BCE) brings significant improvements.
This empirical evidence suggests that \textit{the ability to mine hard patches} contributes to better feature representations.

\begin{table*}[t]
    \centering
    \caption{
    Comparison with state-of-the-art alternatives on \textbf{Something-Something v2}~\cite{goyal2017something}.
    Only the results obtained with similar backbones are listed here. 
    ``IN'' indicates ImageNet~\cite{russakovsky2015imagenet}.
    Results achieved by using a larger backbone are shown in \gc{grayscale}.
    $\dag$ means our reproduced results.
    ``n/a'' indicates the numbers are not available for us.
    }
    \label{tab:sota_SSv2}
    \vspace{-10pt}
    \setlength{\tabcolsep}{7pt}
    \begin{tabular}{ll llc rrcc}
    \toprule
    method & venue & backbone & pre-train & frames & GFLOPs & param. & top-1 & top-5 \\
    \midrule
    \multicolumn{9}{l}{\textit{Supervised Training with Labels}} \\
    SlowFast~\cite{feichtenhofer2019slowfast} & \pub{ICCV'19} & R101~\cite{he2016deep} & K400~\cite{kay2017kinetics} & 8+32 & 106$\times$1$\times$3 & 53 & 63.1 & 87.6 \\
    MViT v1~\cite{fan2021multiscale} & \pub{CVPR'21} & MViT-B~\cite{fan2021multiscale} & K400~\cite{kay2017kinetics} & 64 & 455$\times$1$\times$3 & 37 & 67.7 & 90.9 \\
    TimeSformer~\cite{bertasius2021space} & \pub{ICML'21} & ViT-B~\cite{dosovitskiy2020image} & IN-21K~\cite{russakovsky2015imagenet} & 8 & 196$\times$1$\times$3 & 121 & 59.5 & n/a \\
    MotionFormer~\cite{patrick2021keeping} & \pub{NeurIPS'21} & ViT-B~\cite{dosovitskiy2020image} & IN-21K~\cite{russakovsky2015imagenet} + K400~\cite{kay2017kinetics} & 16 & 370$\times$1$\times$3 & 109 & 66.5 & 90.1 \\
    VideoSwin~\cite{liu2022video} & \pub{CVPR'22} & Swin-B~\cite{liu2021swin} & IN-21K~\cite{russakovsky2015imagenet} + K400~\cite{kay2017kinetics} & 32 & 321$\times$1$\times$3 & 88 & 69.6 & 92.7 \\
    \gc{TimeSformer}~\cite{bertasius2021space} & \pub{ICML'21} & \gc{ViT-L}~\cite{dosovitskiy2020image} & \gc{IN-21K}~\cite{russakovsky2015imagenet} & \gc{64} & \gc{5549$\times$1$\times$3} & \gc{430} & \gc{62.4} & \gc{n/a} \\
    \gc{ViViT FE}~\cite{arnab2021vivit} & \pub{ICCV'21} & \gc{ViT-L}~\cite{dosovitskiy2020image} & \gc{IN-21K}~\cite{russakovsky2015imagenet} \gc{+ K400}~\cite{kay2017kinetics} & \gc{32} & \gc{995$\times$4$\times$3} & \gc{n/a} & \gc{65.9} & \gc{89.9} \\
    \gc{MotionFormer}~\cite{patrick2021keeping} & \pub{NeurIPS'21} & \gc{ViT-L}~\cite{dosovitskiy2020image} & \gc{IN-21K}~\cite{russakovsky2015imagenet} \gc{+ K400}~\cite{kay2017kinetics} & \gc{32} & \gc{1185$\times$1$\times$3} & \gc{382} & \gc{68.1} & \gc{91.2} \\
    \midrule
    \multicolumn{9}{l}{\textit{Self-Supervised Pre-Training without Labels}} \\
    VideoMAE$_{800e}^{\dag}$~\cite{tong2022videomae} & \pub{NeurIPS'22} & ViT-B~\cite{dosovitskiy2020image} & SSv2~\cite{goyal2017something} & 16 & 180$\times$2$\times$3 & 87 & 69.3 & 91.7 \\
    %
    \rowcolor{Light}
    HPM$_{800e}$ & \pub{Ours} & ViT-B~\cite{dosovitskiy2020image} & SSv2~\cite{goyal2017something} & 16 & 180$\times$2$\times$3 & 87 & 69.8 & 92.1 \\
    VideoMAE$_{2400e}$~\cite{tong2022videomae} & \pub{NeurIPS'22} & ViT-B~\cite{dosovitskiy2020image} & SSv2~\cite{goyal2017something} & 16 & 180$\times$2$\times$3 & 87 & 70.8 & 92.4 \\
    %
    \rowcolor{Light}
    HPM$_{2400e}$ & \pub{Ours} & ViT-B~\cite{dosovitskiy2020image} & SSv2~\cite{goyal2017something} & 16 & 180$\times$2$\times$3 & 87 & \textbf{71.3} & \textbf{92.7} \\
    \gc{MaskFeat$_{\uparrow312}$}~\cite{wei2022masked} & \pub{CVPR'22} & \gc{MViT-L}~\cite{fan2021multiscale} & \gc{K600}~\cite{kay2017kinetics} & \gc{40} & \gc{2828$\times$1$\times$3} & \gc{218} & \gc{75.0} & \gc{95.0} \\
    \gc{MAEST$_{1600e}$}~\cite{feichtenhofer2022masked} & \pub{NeurIPS'22} & \gc{ViT-L}~\cite{dosovitskiy2020image} & \gc{K400}~\cite{kay2017kinetics} & \gc{16} & \gc{597$\times$1$\times$3} & \gc{305} & \gc{72.1} & \gc{93.9} \\
    \gc{VideoMAE$_{2400e}$}~\cite{tong2022videomae} & \pub{NeurIPS'22} & \gc{ViT-L}~\cite{dosovitskiy2020image} & \gc{SSv2}~\cite{goyal2017something} & \gc{16} & \gc{597$\times$2$\times$3} & \gc{305} & \gc{74.3} & \gc{94.6} \\
    \bottomrule
    \end{tabular}
\end{table*}

\begin{table*}[t]
    \centering
    \caption{
    Comparison with state-of-the-art alternatives on \textbf{Kinetics-400}~\cite{kay2017kinetics}.
    Only the results obtained with similar backbones are listed here. 
    ``IN'' indicates ImageNet~\cite{russakovsky2015imagenet}.
    Results achieved by using a larger backbone are shown in \gc{grayscale}.
    ``n/a'' indicates the numbers are not available for us.
    }
    \label{tab:sota_k400}
    \vspace{-10pt}
    \setlength{\tabcolsep}{7.5pt}
    \begin{tabular}{ll ll rrrcc}
    \toprule
    method & venue & backbone & pre-train & frames & GFLOPs & param. & top-1 & top-5 \\
    \midrule
    \multicolumn{9}{l}{\textit{Supervised Training with Labels}} \\
    SlowFast~\cite{feichtenhofer2019slowfast} & \pub{ICCV'19} & R101~\cite{he2016deep}+NL~\cite{wang2018non} & & 16+64 & 234$\times$10$\times$3 & 60 & 79.8 & 93.9 \\
    MViT v1~\cite{fan2021multiscale} & \pub{CVPR'21} & MViT-B~\cite{fan2021multiscale} & & 32 & 170$\times$5$\times$1 & 37 & 80.2 & 94.4 \\
    VideoSwin~\cite{liu2022video} & \pub{CVPR'22} & Swin-B~\cite{liu2021swin} & IN-21K~\cite{russakovsky2015imagenet} & 32 & 284$\times$4$\times$3 & 88 & 82.7 & 95.5 \\
    
    \gc{MotionFormer}~\cite{patrick2021keeping} & \pub{NeurIPS'21} & \gc{ViT-L}~\cite{dosovitskiy2020image} & \gc{IN-21K}~\cite{russakovsky2015imagenet} & \gc{32} & \gc{1185$\times$10$\times$3} & \gc{382} & \gc{80.2} & \gc{94.8} \\
    \gc{TimeSformer}~\cite{bertasius2021space} & \pub{ICML'21} & \gc{ViT-L}~\cite{dosovitskiy2020image} & \gc{IN-21K}~\cite{russakovsky2015imagenet} & \gc{96} & \gc{8353$\times$1$\times$3} & \gc{430} & \gc{80.7} & \gc{94.7} \\
    \gc{ViViT FE}~\cite{arnab2021vivit} & \pub{ICCV'21} & \gc{ViT-L}~\cite{dosovitskiy2020image} & \gc{IN-21K}~\cite{russakovsky2015imagenet} & \gc{128} & \gc{3980$\times$1$\times$3} & \gc{n/a} & \gc{81.7} & \gc{93.8} \\
    \midrule
    \multicolumn{9}{l}{\textit{Self-Supervised Pre-Training without Labels}} \\
    VideoMAE$_{800e}$~\cite{tong2022videomae} & \pub{NeurIPS'22} & ViT-B~\cite{dosovitskiy2020image} & K400~\cite{kay2017kinetics} & 16 & 180$\times$5$\times$3 & 87 & 80.0 & 94.4 \\
    
    %
    \rowcolor{Light}
    HPM$_{800e}$ & \pub{Ours} & ViT-B~\cite{dosovitskiy2020image} & K400~\cite{kay2017kinetics} & 16 & 180$\times$5$\times$3 & 87 & 80.8 & 94.6 \\
    MAEST$_{1600e}$~\cite{feichtenhofer2022masked} & \pub{NeurIPS'22} & ViT-B~\cite{dosovitskiy2020image} & K400~\cite{kay2017kinetics} & 16 & 180$\times$7$\times$3 & 87 & 81.3 & 94.9 \\
    VideoMAE$_{1600e}$~\cite{tong2022videomae} & \pub{NeurIPS'22} & ViT-B~\cite{dosovitskiy2020image} & K400~\cite{kay2017kinetics} & 16 & 180$\times$5$\times$3 & 87 & 81.5 & \textbf{95.1} \\
    %
    \rowcolor{Light}
    HPM$_{1600e}$ & \pub{Ours} & ViT-B~\cite{dosovitskiy2020image} & K400~\cite{kay2017kinetics} & 16 & 180$\times$5$\times$3 & 87 & \textbf{81.6} & \textbf{95.1} \\
    \gc{MAEST$_{1600e}$}~\cite{feichtenhofer2022masked} & \pub{NeurIPS'22} & \gc{ViT-L}~\cite{dosovitskiy2020image} & \gc{K400}~\cite{kay2017kinetics} & \gc{16} & \gc{597$\times$7$\times$3} & \gc{305} & \gc{84.8} & \gc{96.2} \\
    \gc{VideoMAE$_{1600e}$}~\cite{tong2022videomae} & \pub{NeurIPS'22} & \gc{ViT-L}~\cite{dosovitskiy2020image} & \gc{K400}~\cite{kay2017kinetics} & \gc{16} & \gc{597$\times$5$\times$3} & \gc{305} & \gc{85.2} & \gc{96.8} \\
    \gc{MaskFeat$_{\uparrow352}$}~\cite{wei2022masked} & \pub{CVPR'22} & \gc{MViT-L}~\cite{fan2021multiscale} & \gc{K600}~\cite{kay2017kinetics} & \gc{40} & \gc{3790$\times$4$\times$3} & \gc{218} & \gc{87.0} & \gc{97.4} \\
    \bottomrule
    \end{tabular}
\end{table*}

\subsubsection{Comparison with Previous Alternatives}

We compare our \method with state-of-the-art approaches on Something-Something v2~\cite{goyal2017something} and Kinetics-400~\cite{kay2017kinetics} in \cref{tab:sota_SSv2} and \cref{tab:sota_k400}, respectively.
We mainly compare methods with similar backbones and computational resources considering a fair comparison.
A wide range of representative alternatives are selected for comparison.

On Something-Something v2, our \method achieves 71.3\% top-1 accuracy when pre-trained with 2400 epochs, outperforming the original VideoMAE~\cite{tong2022videomae} by +0.5\%.
On Kinetics-400, \method outperforms the original VideoMAE~\cite{tong2022videomae} +0.8\% and +0.1\% with 800 and 1600 epochs of pre-training, respectively.
Empirical evidence illustrated in \cref{tab:sota_SSv2} and \cref{tab:sota_k400} demonstrate that solving more challenging proxy tasks generated by \method contributes to better spatiotemporal feature representations.

The small improvements on Kinetics-400 with 1600 epochs of pre-training might be due to the \textit{scene-centric} attribute of Kinetics-400, where motion information is \textit{less} important compared with Something-Something v2.
In fact, it is better to evaluate video pre-training approaches under motion-centric scenarios instead of almost static benchmarks like Kinetics-400.
Because these methods are expected to make the model aggregate the temporal information and extract generalizable representations.

%% file: sections/5.conclusion.tex
\section{Conclusion}\label{sec:conclusion}

In this paper, we find it necessary to \textit{make the model stand in the shoes of a teacher} for MVM pre-training, and verify that the patch-wise reconstruction loss can naturally be the metric of the reconstruction difficulty.
To this end, we propose \method, which introduces an auxiliary reconstruction loss prediction task, and thus guides the training procedure iteratively in a produce-and-solve manner.
We also come up with an easy-to-hard mask generation manner.
Experimentally, \method bootstraps the performance of masked image modeling across various downstream tasks.
Ablations across different learning targets show that \method, as a plug-and-play module, can be effortlessly incorporated into existing frameworks (\textit{e.g.}, pixel regression~\cite{he2022masked, xie2022simmim, tong2022videomae} and feature prediction~\cite{zhou2021ibot, dong2022bootstrapped, wei2022masked}) and bring \textit{consistent} performance improvements.
Moreover, we empirically find that increasing the difficulty of the pretext task while retaining a certain degree of randomness is crucial.
We further extend \method to video pre-training benchmarks and show its effectiveness.

\paragraph{Broader impact.}
Techniques that mine hard examples are widely used in object detection~\cite{lin2017focal, shrivastava2016training, li2019gradient}.
Loss prediction can be a brand-new alternative.
Furthermore, it can be also used as a technique to filter high-quality pseudo-labels in label-efficient learning~\cite{wang2022semi, du2022learning, wang2023balancing, wang2023pulling, wang2023using}.
Meanwhile, as shown in \cref{fig:discriminative} and \cref{fig:coco_half}, \textit{the salient area tends to have a higher predicted loss}, and thus \method may also be used for saliency detection~\cite{wang2021salient} and unsupervised segmentation~\cite{van2021unsupervised}.
We hope these perspectives will inspire future work.
